\documentclass{article}

\usepackage[table]{xcolor} 

\usepackage[preprint]{neurips_data_2022}

\usepackage[utf8]{inputenc} 
\usepackage[T1]{fontenc}    
\usepackage{hyperref}       
\usepackage{url}            
\usepackage{booktabs}       
\usepackage{amsfonts}       
\usepackage{nicefrac}       
\usepackage{microtype}      
\usepackage{xcolor}         
\usepackage{graphicx}
\usepackage{caption}
\usepackage{subcaption}
\usepackage{enumitem}
\usepackage{csvsimple}

\usepackage[textsize=tiny]{todonotes}

\usepackage{ulem}

\title{Why do tree-based models still outperform deep learning on tabular data?}

\author{%
  Léo Grinsztajn \\
  Soda, Inria Saclay\\
  \texttt{leo.grinsztajn@inria.fr} \\
  \And
   Edouard Oyallon \\
   ISIR, CNRS, Sorbonne University
  \And
  Gaël Varoquaux \\
  Soda, Inria Saclay
}

\begin{document}

\maketitle

\begin{abstract}
While deep learning has enabled tremendous progress on text and image datasets, its superiority on tabular data is not clear. We contribute extensive benchmarks of standard and novel deep learning methods as well as tree-based models such as XGBoost and Random Forests, across a large number of datasets and hyperparameter combinations.
We define a standard set of 45 datasets from varied domains with clear characteristics of tabular data and a benchmarking methodology accounting for both fitting models and finding good hyperparameters.
Results show that tree-based models remain state-of-the-art on medium-sized data ($\sim$10K samples) even without accounting for their superior speed. To understand this gap, we conduct an empirical investigation into the differing inductive biases of tree-based models and Neural Networks (NNs). This leads to a series of challenges which should guide researchers aiming to build tabular-specific NNs: \textbf{1.} be robust to uninformative features, \textbf{2.} preserve the orientation of the data, and \textbf{3.} be able to easily learn irregular functions. To stimulate research on tabular architectures, we contribute a standard benchmark and raw data for baselines: every point of a 20\,000 compute hours hyperparameter search for each learner.
\end{abstract}

\section{Introduction}


Deep learning has enabled tremendous progress for learning on image, language, or even audio datasets. On tabular data, however, the picture is muddier and ensemble models based on decision trees like XGBoost remain the go-to tool for most practitioners \citep{StateDataScience} and data science competitions \citep{kossenSelfAttentionDatapointsGoing2021}. Indeed deep learning architectures have been crafted to create inductive biases matching invariances and spatial dependencies of the data. Finding corresponding invariances is hard in tabular data, made of heterogeneous features, small sample sizes, extreme values.

Creating tabular-specific deep learning architectures is a very active area of research (see \autoref{sec:related}) given that
tree-based models are not differentiable, and thus cannot be easily composed and jointly trained with other deep learning blocks. Most corresponding publications claim to beat or match tree-based models, but their claims have been put into question: a simple Resnet seems to be competitive with some of these new models \citep{gorishniyRevisitingDeepLearning2021}, and most of these methods seem to fail on new datasets \citep{shwartz-zivTabularDataDeep2021}. 
Indeed, the lack of an established benchmark for tabular data learning 
provides additional degrees of freedom to researchers when evaluating their method.
Furthermore, most tabular datasets available online are small compared to benchmarks in other machine learning subdomains, such as ImageNet~\citep{ImageNetLargescaleHierarchical}, making evaluation noisier. These issues add up to other sources of unreplicability across machine learning, such as unequal hyperparameters tuning efforts \citep{liptonTroublingTrendsMachine2019} or failure to account for statistical uncertainty in benchmarks \citep{bouthillierAccountingVarianceMachine2021a}. To alleviate these concerns, we contribute a tabular data benchmark with a precise methodology for datasets inclusion and hyperparameter tuning. This enables us to evaluate recent deep learning models which have not yet been independently evaluated, and to show that tree-based models remain state-of-the-art on medium-sized tabular datasets, even without accounting for the slower training of deep learning algorithms.

Impressed by the superiority of tree-based models on tabular data, we strive to understand which \textit{inductive biases} make them well-suited for these data. By transforming tabular datasets to modify the performances of different models, we uncover differing biases of tree-based models and deep learning algorithms which partly explain their different performances: neural networks struggle to learn irregular patterns of the target function, and their rotation invariance hurt their performance, in particular when handling the numerous uninformative features present in tabular data.





Our contributions are as follow:  \textbf{1.} We create a new benchmark for tabular data, with a precise methodology for choosing and preprocessing a large number of representative datasets. We share these datasets through OpenML \citep{vanschorenOpenMLNetworkedScience2014}, which makes them easy to use. \textbf{2.} We extensively compare deep learning models and tree-based models on generic tabular datasets in multiple settings, accounting for the cost of choosing hyperparameters. We also share the raw results of our random searches, which will enable researchers to cheaply test new algorithms for a fixed hyperparameter optimization budget.
\textbf{3.} We investigate empirically why tree-based models outperform deep learning, by finding data transformations which narrow or widen their performance gap. 
This highlights desirable biases for tabular data learning, which we hope will help other researchers to successfully build deep learning models for tabular data.

In Sec.\,\ref{sec:related} we cover related work.
Sec. \ref{bench} gives a short description of our benchmark methodology, including datasets, data processing, and  hyper-parameter tuning. Then, Sec. \ref{numeric} shows our raw results on deep learning and tree-based models after an extensive random search. Finally, Sec. \ref{ablation} provides the results of an empirical study which exhibit desirable implicit biases of tabular datasets. 

\section{Related work}

\label{sec:related}%

\paragraph{Deep learning for tabular data} As described by \cite{borisovDeepNeuralNetworks2021} in their review of the field, there have been various attempts to make deep learning work on tabular data: data encoding techniques to make tabular data better suited for deep learning \citep{hancockSurveyCategoricalData2020, yoonVIMEExtendingSuccess2020}, "hybrid methods" to benefit from the flexibility of NNs while keeping the inductive biases of other algorithms like tree-based models \citep{layRandomHingeForest2018, popovNeuralObliviousDecision2020,abutbulDNFNetNeuralArchitecture2020,hehnEndtoEndLearningDecision2019, tannoAdaptiveNeuralTrees2019,chenAttentionAugmentedDifferentiable2020, kontschiederDeepNeuralDecision2015, rodriguezInterpretableReinforcementLearning2019, popovNeuralObliviousDecision2020, layRandomHingeForest2018} or Factorization Machines \cite{guoDeepFMFactorizationMachineBased2017}, tabular-specific transformers architectures \cite{somepalliSAINTImprovedNeural2021,kossenSelfAttentionDatapointsGoing2021,arikTabNetAttentiveInterpretable2019,huangTabTransformerTabularData2020}, and various regularization techniques to adapt classical architectures to tabular data \citep{louniciMuddlingLabelRegularization2021, shavittRegularizationLearningNetworks2018,kadraWelltunedSimpleNets2021, fiedlerSimpleModificationsImprove2021}. In this paper, we focus on architectures directly inspired by classic deep learning models, in particular Transformers and Multi-Layer-Perceptrons (MLPs).

\paragraph{Comparisons between NNs and tree-based models} The most comprehensive comparisons of machine learning algorithms have been published before the advent of new deep learning methods, or on specific problems \citep{fernandez-delgadoWeNeedHundreds2014, sakrComparisonMachineLearning2017, korotcovComparisonDeepLearning2017,uddinComparingDifferentSupervised2019}. Recently, \cite{shwartz-zivTabularDataDeep2021} evaluated modern tabular-specific deep learning methods, but their goal was more to reveal that "New deep learning architectures fail to generalize to new datasets" than to create a comprehensive benchmark. \cite{borisovDeepNeuralNetworks2022} benchmarked recent algorithms in their review of deep learning for tabular data, but only on 3 datasets, and "highlight[ed]
the need for unified benchmarks" for tabular data. Most papers introducing a new architecture for tabular data benchmark various algorithms, but with a highly variable evaluation methodology, a small number of datasets, and the evaluation can be biased toward the authors' model \cite{shwartz-zivTabularDataDeep2021}. The paper closest to our work is \cite{gorishniyRevisitingDeepLearning2021}, benchmarking novel algorithms, on 11 tabular datasets. We provide a more comprehensive benchmark, with 45 datasets, split across different settings (medium-sized / large-size, with/without categorical features), accounting for the hyperparameter tuning cost, to establish a standard benchmark.


\paragraph{No standard benchmark for tabular data} Unlike other machine learning subfields such as computer vision \citep{ImageNetLargescaleHierarchical} or NLP  \citep{wangSuperGLUEStickierBenchmark2020}, there are no standard benchmarks for tabular data. There exist generic machine learning benchmarks, but, to the our knowledge, none are specific to tabular data. For instance, OpenML benchmarks CC-18, CC-100, \citep{bischlOpenMLBenchmarkingSuites2021} and AutoML Benchmark \cite{gijsbersOpenSourceAutoML2019} contain tabular data, but also include images and artificial datasets, which may explain why they have not been used in tabular deep learning papers.

\paragraph{Understanding the difference between NNs and tree-based models} To our knowledge, this is the first empirical investigation of \textsl{why} tree-based models outperform NNs on tabular data. Some speculative explanations, however, have been offered \citep{klambauerSelfNormalizingNeuralNetworks2017,borisovDeepNeuralNetworks2021}. \cite{kadraWelltunedSimpleNets2021} claims that searching across  13 regularization techniques for MLPs to find a dataset-specific combination gives state-of-the-art performances. This provides a partial explanation: MLPs are expressive enough for tabular data but may suffer from a lack of proper regularization.

\section{A benchmark for tabular learning}\label{bench}
\subsection{45 reference tabular datasets}%

For our benchmark, we compiled 45 tabular datasets from various domains provided mainly by OpenML, listed in \ref{supp:datasets} and selected via the following criteria:
\begin{description}[topsep=0pt,itemsep=0ex,partopsep=0ex,parsep=0ex,leftmargin=2ex]
    \item[Heterogeneous columns.] Columns should correspond to features of different nature. This excludes images or signal datasets where each column corresponds to the same signal on different sensors.
    \item[Not high dimensional.] We only keep datasets with a $d/n$ ratio below 1/10.
    \item[Undocumented datasets] We remove datasets where too little information is available. We did keep datasets with hidden column names if it was clear that the features were heterogeneous.
    \item[I.I.D. data.] We remove stream-like datasets or time series.
    \item[Real-world data.] We remove artificial datasets but keep some simulated datasets. The difference is subtle, but we try to keep simulated datasets if learning these datasets are of practical importance (like the Higgs dataset), and not just a toy example to test specific model capabilities.
    \item[Not too small.] We remove datasets with too few features (< 4) and too few samples (< 3\,000). For benchmarks on numerical features only, we remove categorical features before checking if enough features and samples are remaining.
    \item[Not too easy.] We remove datasets which are too easy. Specifically, we remove a dataset if a default Logistic Regression (or Linear Regression for regression) reach a score whose relative difference with the score of both a default Resnet (from \cite{gorishniyRevisitingDeepLearning2021}) and a default HistGradientBoosting model (from scikit learn) is below 5\%. Other benchmarks use different metrics to remove too easy datasets, like removing datasets which can be learnt perfectly by a single decision classifier \citep{bischlOpenMLBenchmarkingSuites2021}, but this does not account for different Bayes rate of different datasets. As tree-based methods have been shown to be superior to Logistic Regression \citep{fernandez-delgadoWeNeedHundreds2014} in our setting, a close score for these two types of models indicates that we might already be close to the best achievable score.
    \item[Not deterministic.] We remove datasets where the target is a deterministic function of the data. This mostly means removing datasets on games like poker and chess. Indeed, we believe that these datasets are very different from most real-world tabular datasets, and should be studied separately. 
\end{description}

\subsection{Removing side issues}

To keep learning tasks as
homogeneous as possible and focus on challenges specific to tabular data, we exclude subproblems which would deserve their own analysis:

\begin{description}[topsep=0pt,itemsep=0ex,partopsep=0ex,parsep=0ex,leftmargin=2ex]
\item[Medium-sized training set] We truncate the training set to 10,000 samples for bigger datasets. This allows us to investigate the medium-sized dataset regime. We study the large-sized (50,000) regime, for which fewer datasets matching our criteria are available, in \ref{supp:benchmarks}.

\item[No missing data] We remove all missing data from the datasets. Indeed, there are numerous techniques for handling
missing data both for tree-based models and NNs, with varying
performances
\citep{perez-lebelBenchmarkingMissingvaluesApproaches2022}. In practice, we first remove columns containing many missing data, then all rows containing at least one missing entry.
\item[Balanced classes] For classification, the target is binarised if there are several classes, by taking the two most numerous classes, and we keep half of samples in each class.
\item[Low cardinality categorical features] We remove categorical features with more than 20 items.
\item[High cardinality numerical features] We remove numerical features with less than 10 unique values. Numerical features with 2 unique values are converted to categorical features.
\end{description}

\subsection{A procedure to benchmark models with hyperparameter selection}\label{hyperparameters}

Hyperparameter tuning leads to uncontrolled variance on a benchmark \citep{bouthillierAccountingVarianceMachine2021a}, especially with a small budget of model evaluations. We design a benchmarking procedure that jointly samples the variance of hyperparameter tuning and explores increasingly high budgets of model evaluations. It relies on random searches for hyper-parameter tuning \citep{hyperopt}.
We use hyperparameter search spaces from the Hyperopt-Sklearn \cite{komerHyperoptSklearnAutomaticHyperparameter2014} when available, from the original paper when possible, and from \cite{gorishniyRevisitingDeepLearning2021} for MLP, Resnet and XGBoost (see \ref{supp:benchmark_details}).
We run a random search of $\approx 400$ iterations per dataset, on CPU for tree-based models and GPU for NNs (more details in \ref{supp:benchmark_details}).

To study performance as a function of the number $n$ of random search iterations, we compute the best hyperparameter combination on the validation set on these $n$ iterations (for each model and dataset), and evaluate it on the test set. We do this 15 times while shuffling the random search order at each time. This gives us bootstrap-like estimates of the expected test score of the best (on the validation set) model after each number of random search iterations. In addition, we always start the random searches with the default hyperparameters of each model.

\paragraph{Resuable code and benchmark raw data}
The code used for all the experiments and comparisons is available at \href{https://github.com/LeoGrin/tabular-benchmark}{https://github.com/LeoGrin/tabular-benchmark}. To help researchers to cheaply add their own algorithms  to the results, we also share at the same link a data table containing results for all iterations of our 20,000 compute-hour random searches.

\subsection{Aggregating results across datasets}

We use the test set accuracy (classification) and R2 score (regression) to measure model performance. To aggregate results across datasets of varying difficulty, we use a metric similar to the distance to the minimum (or average distance to the minimum (ADTM) when averaged across datasets), used in \cite{feurerAutoSklearnHandsfreeAutoML2021} and introduced in \cite{wistubaLearningHyperparameterOptimization2015}. This metric consists in normalizing each test accuracy between 0 and 1 via an affine renormalization between the top-performing  and worse-performing models. Instead of the worse-performing model, we use models achieving the 10\% (classification) or 50\% (regression) test error quantile. Indeed, the worse scores are achieved by outlier models and are not representative of the difficulty of the dataset. For regression tasks, we clip all negative scores (i.e below 50\% scores) to 0 to reduce the influence of very low scores.


\subsection{Data preparation}

We strive for as little manual preprocessing as possible, applying only the following transformations:
\begin{description}[topsep=0pt,itemsep=0ex,partopsep=0ex,parsep=0ex,leftmargin=2ex]
\item[Gaussianized features] For NN training,  the features are
Gaussianized with Scikit-learn's \texttt{QuantileTransformer}.

\item[Transformed regression targets] In regression settings, target variables are log-transformed when their distributions are heavy-tailed (e.g house prices, see \ref{supp:datasets}). In addition, we add as an hyperparameter the possibility to Gaussienize the target variable for model fit, and transform it back for evaluation (via ScikitLearn's  TransformedTargetRegressor and QuantileTransformer).

\item[OneHotEncoder] For models which do not handle categorical variables natively, we encode categorical features using ScikitLearn's OneHotEncoder.
\end{description}

\section{Tree-based models still outperform deep learning on tabular data.}\label{numeric}

\subsection{Models benchmarked}
For tree-based models, we choose 3 state-of-the-art models used by practitioners: Scikit Learn’s RandomForest, GradientBoostingTrees (GBTs) (or HistGradientBoostingTrees when using categorical features), and XGBoost \cite{chenXGBoostScalableTree2016a}. We benchmark the following deep models:
\begin{description}[topsep=0pt,leftmargin=3ex,itemsep=0ex]
    \item[MLP]: a classical MLP from
\cite{gorishniyRevisitingDeepLearning2021}. The only improvement beyond a
simple MLP is using Pytorch's {\tt ReduceOnPlateau}
learning rate scheduler.
    \item[Resnet]: as in
\cite{gorishniyRevisitingDeepLearning2021}, similar to \textbf{MLP} with dropout, batch/layer normalization, and skip connections.
    \item[FT\_Transformer]: a simple Transformer model combined with a module embedding categorical and numerical features, created in \cite{gorishniyRevisitingDeepLearning2021}. We choose this model because it was benchmarked in a convincing way against tree-based models and other tabular-specific models. It can thus be considered a “best case” for Deep learning models on tabular data.
    \item[SAINT]: a Transformer model with an embedding module and an inter-samples attention mechanism, proposed in
\cite{somepalliSAINTImprovedNeural2021}. We include this model because it was the best performing deep model in \cite{borisovDeepNeuralNetworks2021}, and to investigate the impact of inter-sample attention, which performs well on tabular data according to \cite{kossenSelfAttentionDatapointsGoing2022a}.
\end{description}

\subsection{Results}\label{results}

\begin{figure}[tb!]
\begin{minipage}{.5\textwidth}
  \includegraphics[width=\linewidth]{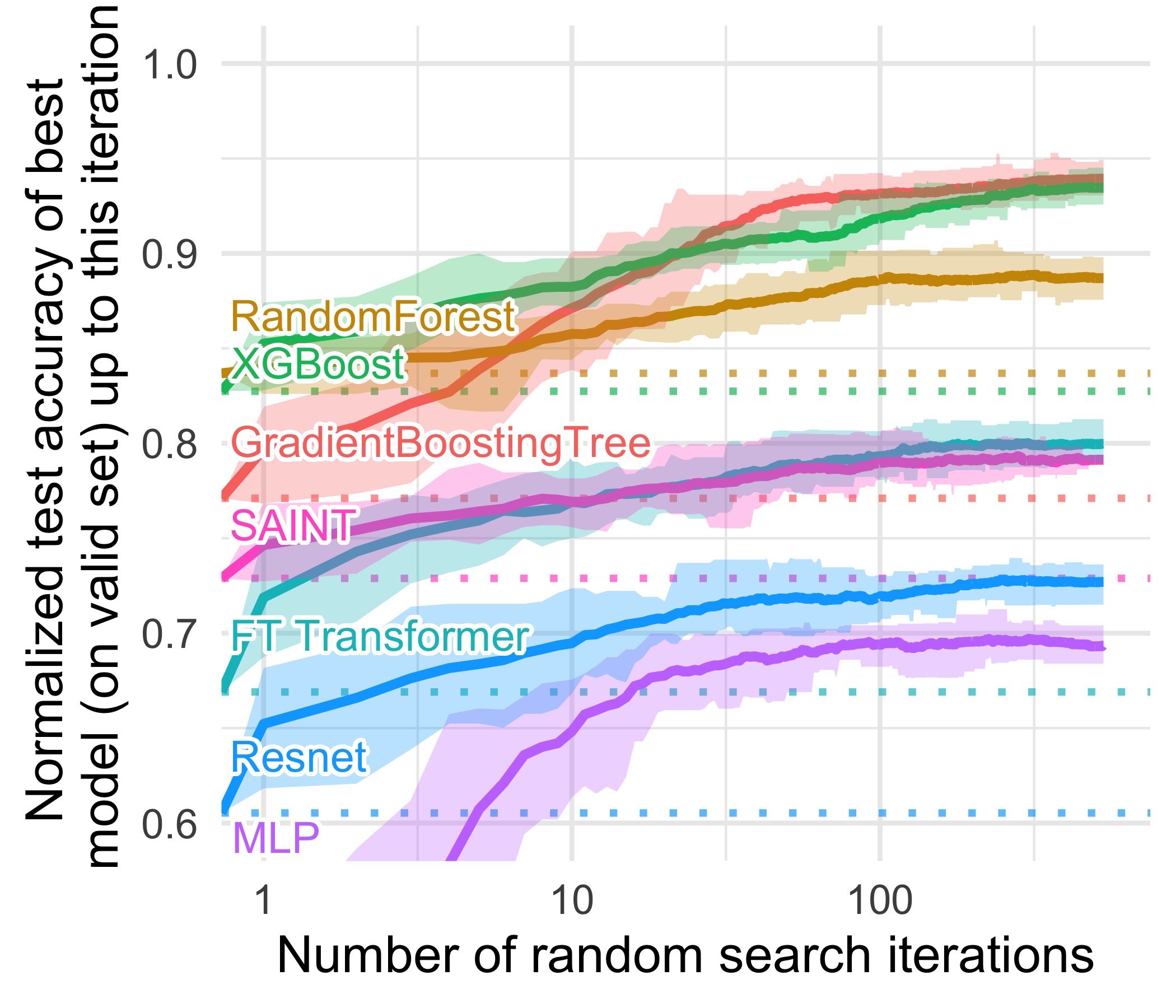}%
  \llap{\raisebox{.8\linewidth}{\parbox{.8\linewidth}{\sffamily{\bfseries Classification} (15 datasets)}}}%
\end{minipage}%
\begin{minipage}{.5\textwidth}
  \includegraphics[width=\linewidth]{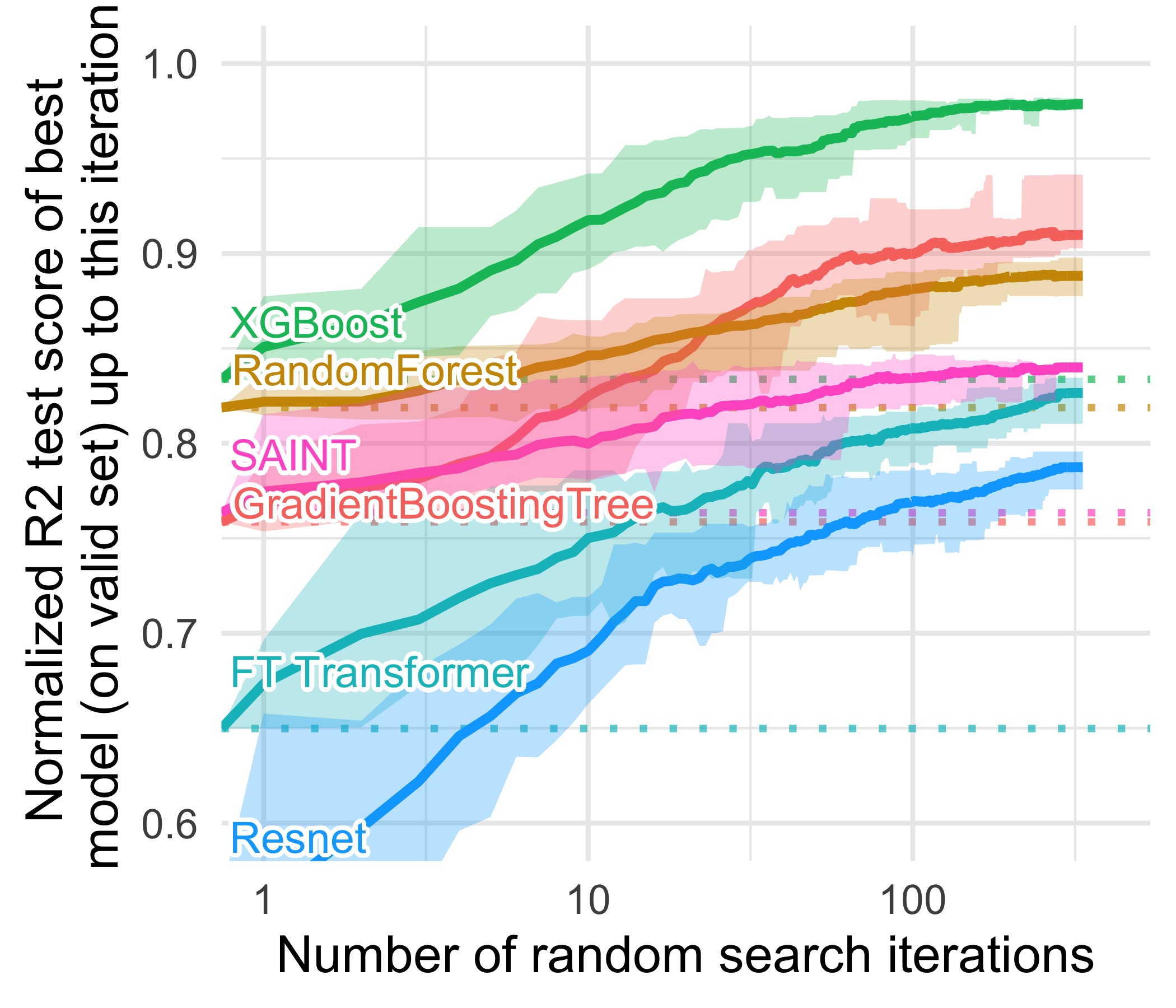}%
    \llap{\raisebox{.8\linewidth}{\parbox{.8\linewidth}{\sffamily{\bfseries Regression} (19 datasets)}}}%
\end{minipage}
\caption{\textbf{Benchmark on medium-sized datasets, with only numerical features}. Dotted lines correspond to the score of the default hyperparameters, which is also the first random search iteration. Each value corresponds to the test score of the best model (on the validation set) after a specific number of random search iterations, averaged on 15 shuffles of the random search order. The ribbon corresponds to the minimum and maximum scores on these 15 shuffles.%
\label{fig:benchmark_numeric}}

\bigskip

\begin{minipage}{.5\textwidth}
  \includegraphics[width=\linewidth]{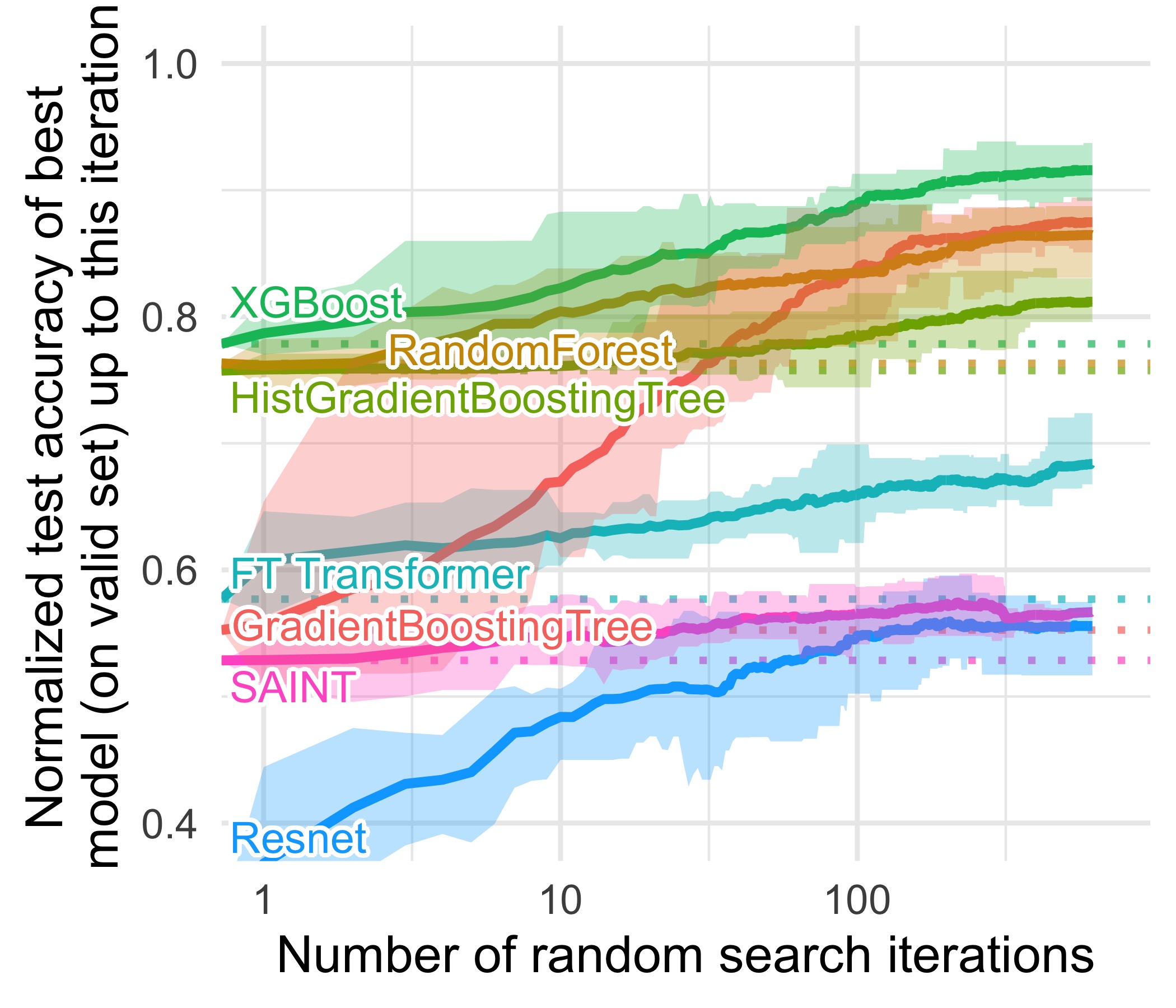}%
    \llap{\raisebox{.82\linewidth}{\parbox{.8\linewidth}{\sffamily{\bfseries Classification} (7 datasets)}}}%
\end{minipage}%
\begin{minipage}{.5\textwidth}
  \includegraphics[width=\linewidth]{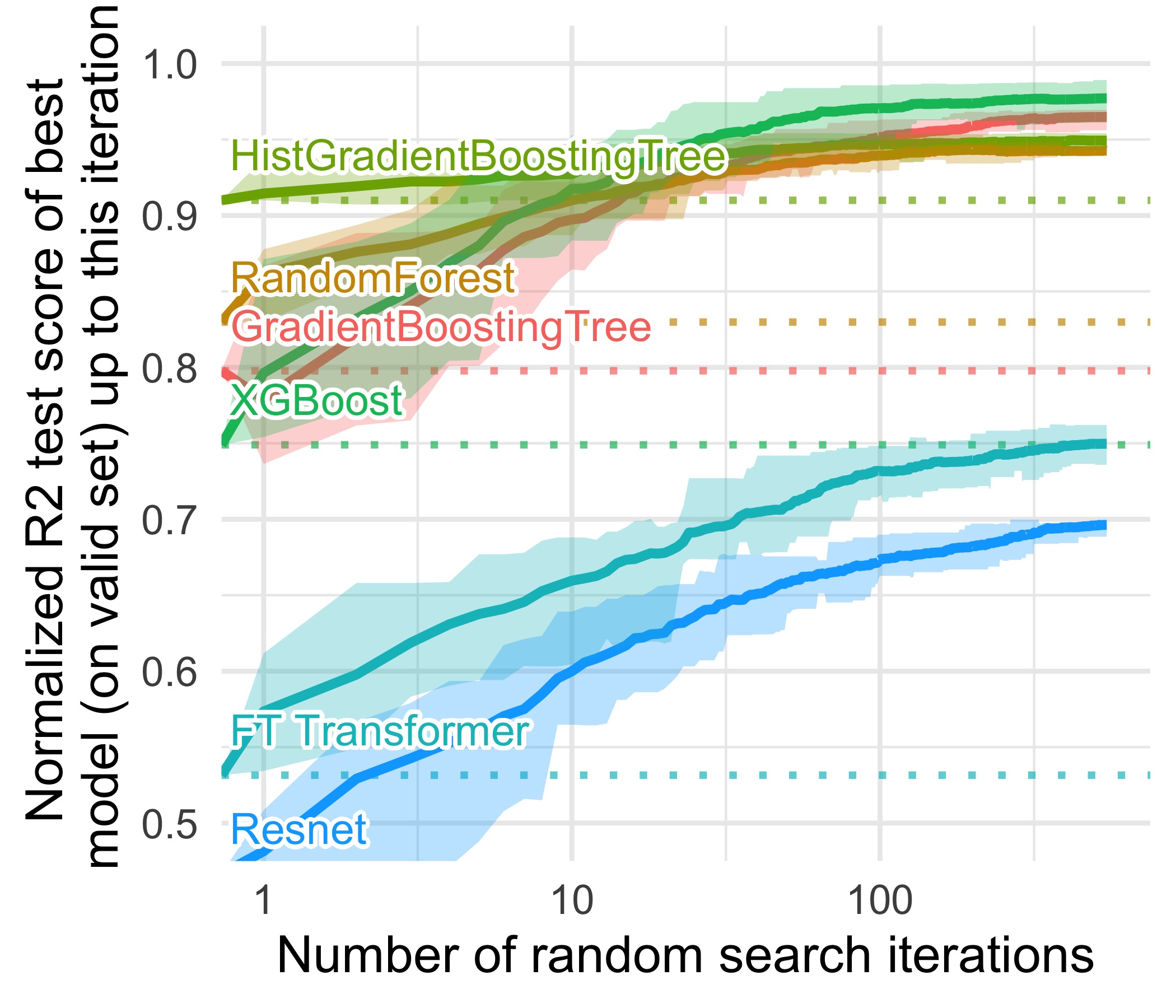}%
      \llap{\raisebox{.82\linewidth}{\parbox{.8\linewidth}{\sffamily{\bfseries Regression} (14 datasets)}}}%
\end{minipage}
\caption{\textbf{Benchmark on medium-sized datasets, with both numerical and categorical features}. Dotted lines correspond to the score of the default hyperparameters, which is also the first random search iteration. Each value corresponds to the test score of the best model (on the validation set) after a specific number of random search iterations, averaged on 15 shuffles of the random search order. The ribbon corresponds to the minimum and maximum scores on these 15 shuffles.}
\label{fig:benchmark_categorical}
\end{figure}

Fig. \ref{fig:benchmark_numeric} and \ref{fig:benchmark_categorical} give benchmark results for different types of datasets (appendix \ref{supp:benchmarks} gives results as a function of computation \textit{time}).

\paragraph{Tuning hyperparameters does not make NNs state-of-the-art}
Tree-based models are superior for every random search budget, and the performance gap stays wide even after a large number of random search iterations. This does not take into account that each random search iteration is generally slower for NNs than for tree-based models (see \ref{supp:benchmarks}).

\paragraph{Categorical variables are not the main weakness of NNs}
Categorical variables are often seen as a major problem for using NNs on tabular data \citep{borisovDeepNeuralNetworks2021}. Our results on numerical variables only do reveal a narrower gap between tree-based models and NNs than including categorical variables. Still, most of this gap subsists when learning on numerical features only.

\section{Empirical investigation: \textit{why} do tree-based models still outperform deep
learning on tabular data?}\label{ablation}

\subsection{Methodology: uncovering inductive biases}

We have seen in Sec. \ref{results} that tree-based models beat NNs across a wide range of hyperparameter choices. This hints to inherent properties of these models which explains their performances on tabular data. Indeed, the best methods on tabular data share two attributes: they are \textbf{ensemble methods}, bagging (Random Forest) or boosting (XGBoost, GBTs), and the weak learner used in these ensembles is a \textbf{decision tree}. The decisive point seems to be the tree aspect:
other boosting and bagging methods with different weak learners exist but are not commonly used for tabular data. In this section, we try to understand the \textit{inductive biases} of decision trees that make them well-suited for tabular data, and how they differ from the inductive biases of NNs. This is equivalent to saying the reverse: which features of tabular data make this type of data easy to learn with tree-based methods yet more difficult with a NN?

To this aim, we apply various transformations to tabular datasets which either narrow or widen the generalization performance gap between NNs and tree-based models, and thus help us emphasize their different inductive biases.
For the sake of simplicity, we restrict our analysis to numerical variables and classification tasks on medium-sized datasets. Results are presented aggregated across datasets, and dataset-specific results are available in \ref{supp:xp_details}, along with additional details on our experiments.



\subsection{Finding 1: NNs are biased to overly smooth solutions}\label{finding_1}






\begin{figure}[t!]
    \begin{minipage}{.55\linewidth}
    \includegraphics[width=\textwidth]{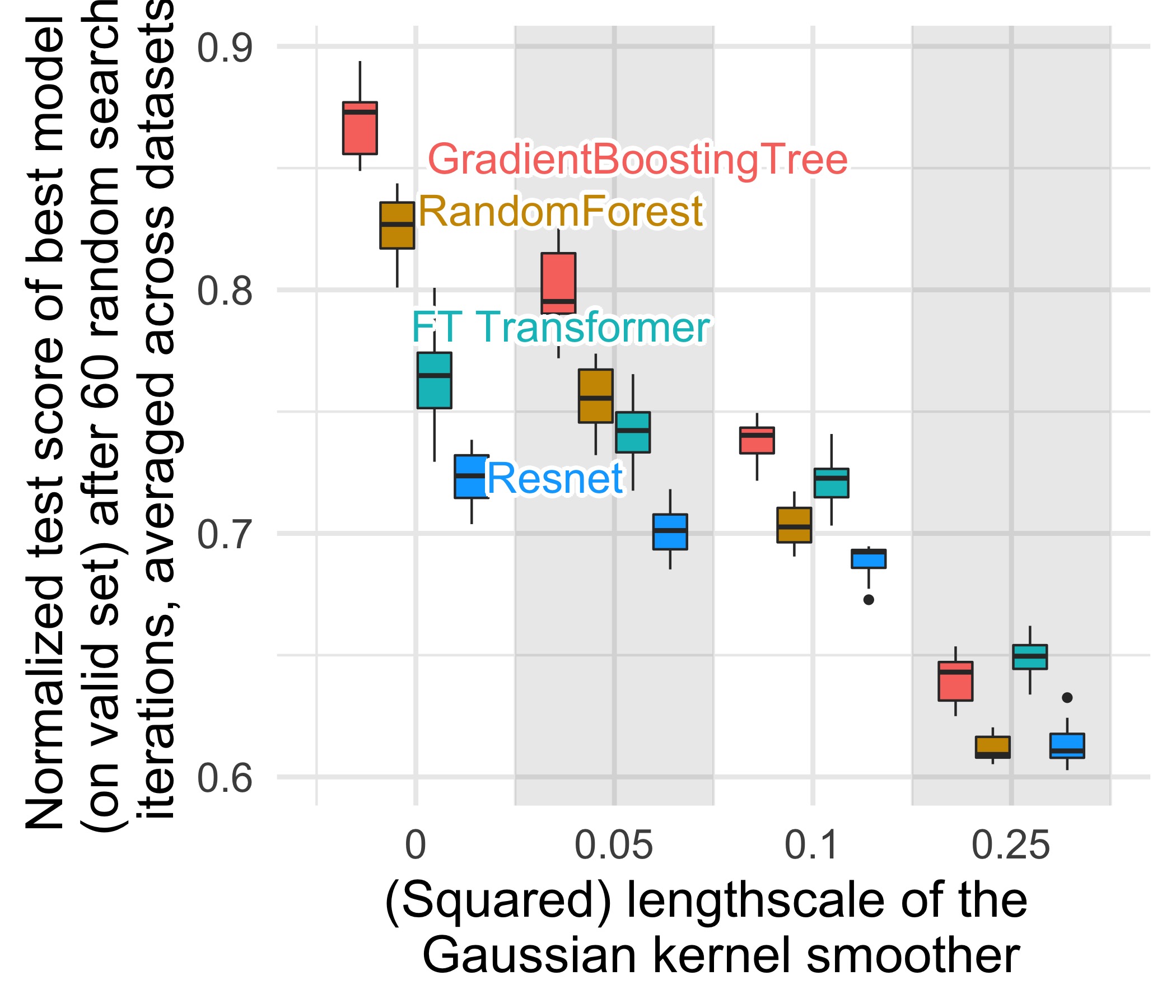}
    \end{minipage}%
        \hfill%
    \begin{minipage}{.42\linewidth}
    \caption{\textbf{Normalized test accuracy of different models for varying smoothing of the target function on the train set}. We smooth the target function through a Gaussian Kernel smoother, whose covariance matrix is the data covariance, multiplied by the (squared) lengthscale of the Gaussian kernel smoother. A lengthscale of 0 corresponds to no smoothing (the original data). All features have been Gaussienized before the smoothing through ScikitLearn's QuantileTransformer. The boxplots represent the distribution of normalized accuracies across 15 re-orderings of the random search.}
    \label{fig:high_frequencies}
    \end{minipage}%
\end{figure}

We transform each \textsl{train} set by smoothing the output with a Gaussian Kernel smoother for varying length-scale values of the kernel (more details are available in \ref{supp:xp_details}). This effectively prevents models from learning irregular patterns of the target function. Fig. \ref{fig:high_frequencies} shows model performance as a function of the length-scale of the smoothing kernel. 
For small lengthscales, smoothing the target function on the train set decreases markedly the accuracy of tree-based models, but barely impacts that of 
NNs.


Such results suggest that the target functions in our datasets are not smooth, and that NNs struggle to fit these irregular functions compared to tree-based models. This is in line with \cite{rahamanSpectralBiasNeural2019}, which finds that NNs are biased toward low-frequency functions. 
Models based on decision trees, which learn piece-wise constant functions, do not exhibit such a bias. 
Our findings do not contradict  papers claiming benefits  from regularization for tabular data \citep{shavittRegularizationLearningNetworks2018,borisovDeepNeuralNetworks2021,kadraRegularizationAllYou2021,louniciMuddlingLabelRegularization2021}, as adequate regularization and careful optimization may allow NNs to learn irregular patterns.
 In \ref{supp:xp_details}, we show some examples of non-smooth patterns which neural networks fail to learn, both in toy and real-world settings.

Note also that our observation could also explain the benefits of the ExU activation used in the Neural-GAM paper \citep{agarwalNeuralAdditiveModels2021}, and of the embeddings used in \cite{gorishniyEmbeddingsNumericalFeatures2022}: the periodic embedding might help the model to learn the high-frequency part of the target function, and the target-aware binning might make the target function smoother.



\subsection{Finding 2: Uninformative features affect more MLP-like NNs}


\paragraph{Tabular datasets contain many uninformative features}

\begin{figure}[tb!]
    \begin{minipage}{.55\linewidth}
    \includegraphics[width=\textwidth]{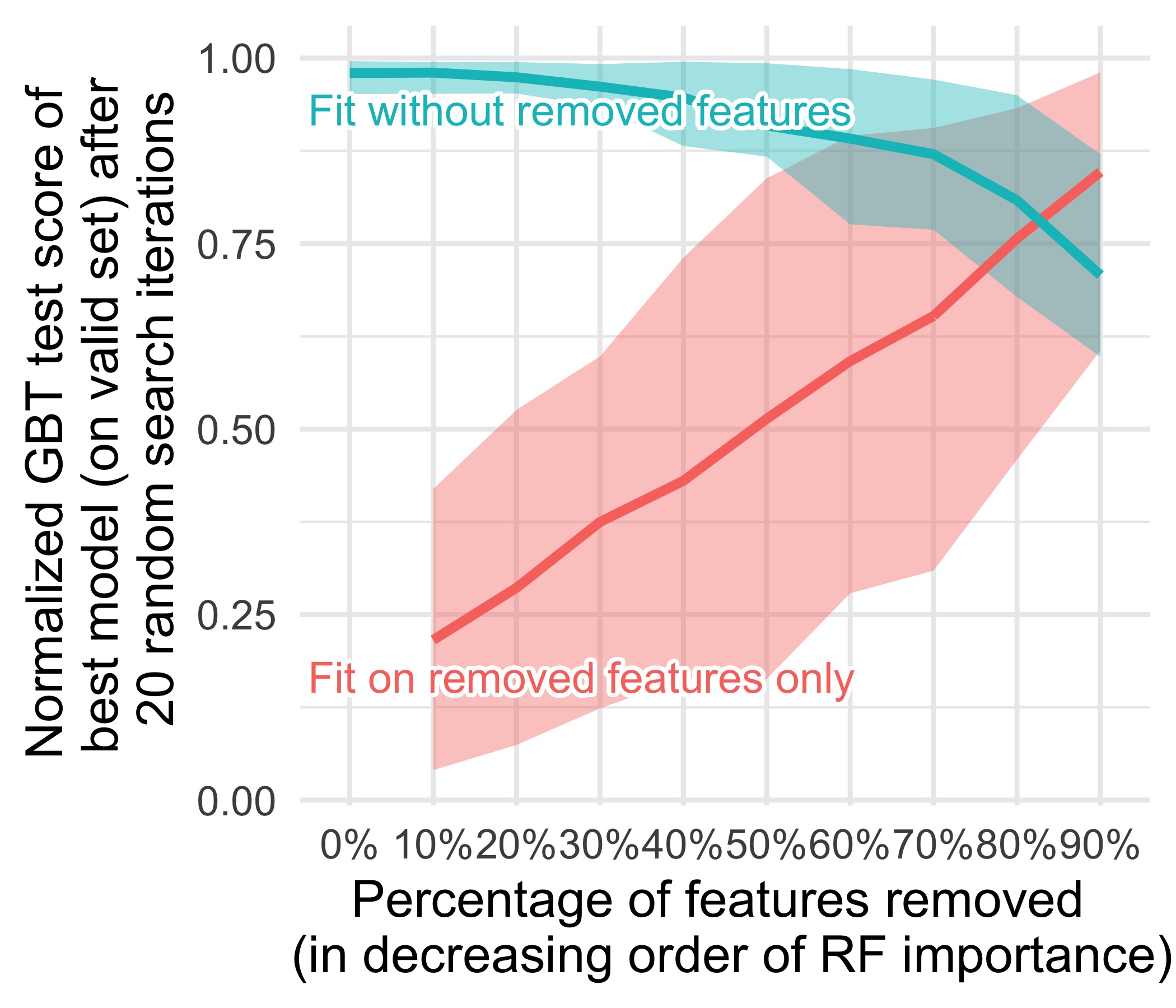}
    \end{minipage}%
    \hfill%
    \begin{minipage}{.43\linewidth}
    \caption{\textbf{Test accuracy of a GBT for varying proportions of removed features}, on our classification benchmark on numerical features. Features are removed in increasing order of feature importance (computed with a Random Forest), and the two lines correspond to the accuracy using the (most important) kept features (blue) or the (least important) removed features (red). A score of 1.0 corresponds to the best score across all models and hyperparameters on each dataset, and 0.0 correspond to random chance. These scores are averaged across 30 random search orders, and the ribbons correspond to the 80\% interval among the different datasets.}
    \label{fig:removed_features}
    \end{minipage}%

\bigskip
\begin{minipage}{.5\textwidth}
    \includegraphics[width=\textwidth]{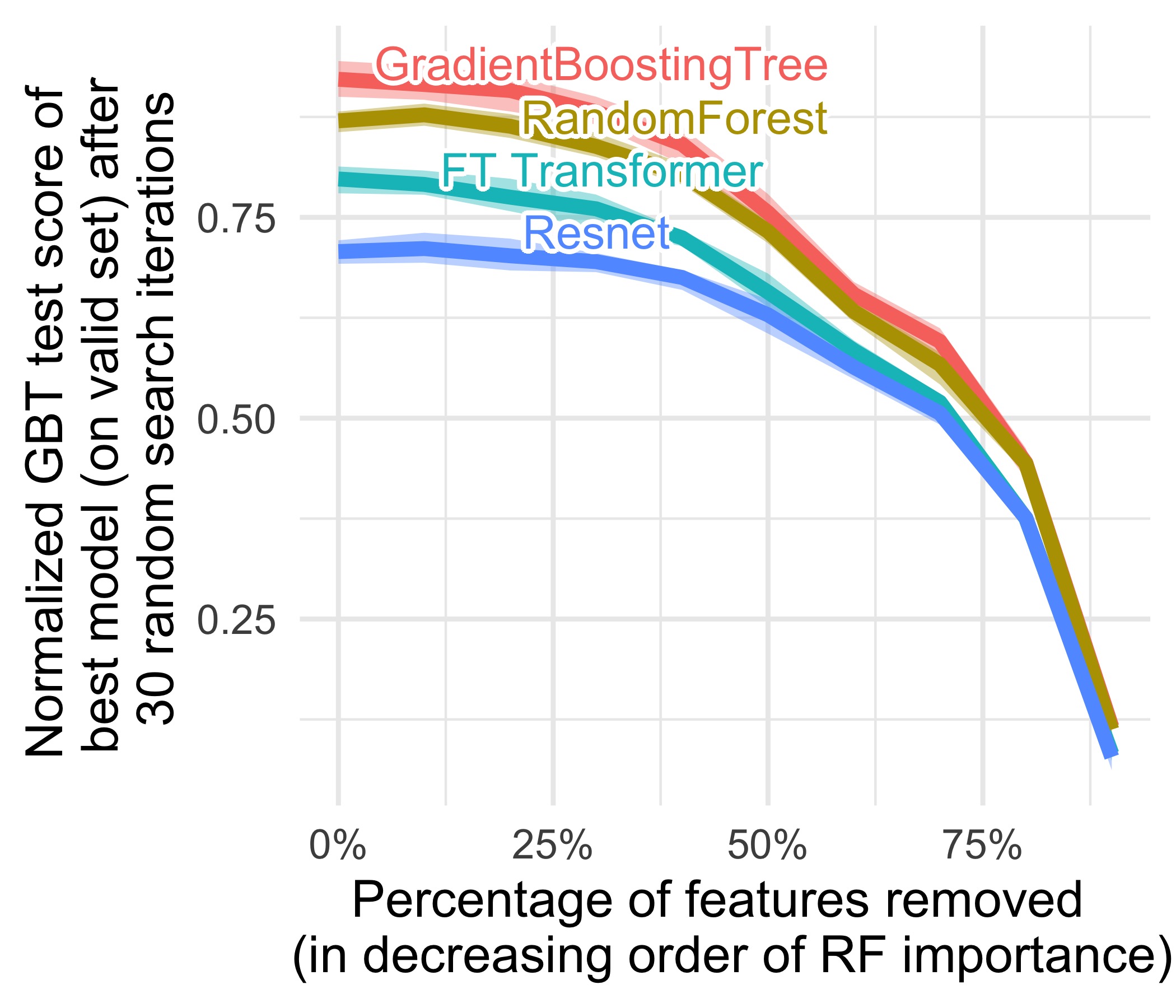}%
        \llap{\raisebox{.25\linewidth}{\parbox{.7\linewidth}{\sffamily\small{\textbf a.} Removing features}}}%
    \label{fig:remove_useless_features}
\end{minipage}%
\begin{minipage}{.5\textwidth}
  \includegraphics[width=\linewidth]{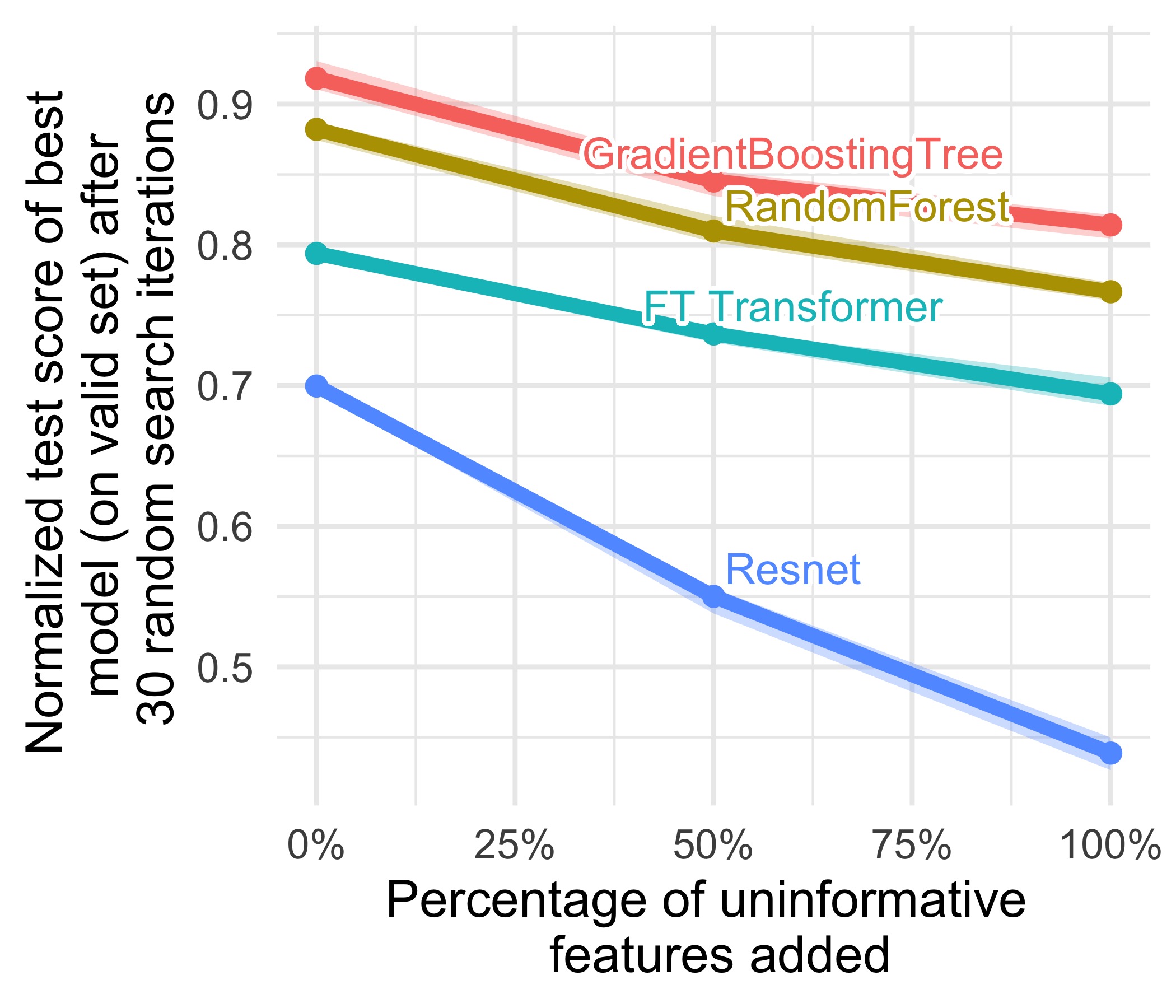}%
        \llap{\raisebox{.25\linewidth}{\parbox{.7\linewidth}{\sffamily\small{\textbf b.} Adding features}}}%
  \label{fig:add_useless_features}
\end{minipage}
\caption{\textbf{Test accuracy changes when removing (a) or adding (b) uninformative features}.  Features
are removed in increasing order of feature
importance (computed with a Random Forest). Added features are sampled from standard Gaussians uncorrelated with the target and with other features. Scores
are averaged across datasets, and the ribbons correspond to the minimum and maximum score among the 30 different random search reorders (starting with the default models).%
\label{fig:features}}%
\end{figure}

For each dataset, we drop an increasingly large fraction of features, according to feature importance (ranked by a Random Forest). Fig. \ref{fig:removed_features} shows that the classification accuracy of a GBT is not much affected by removing up to half of the features.

Furthermore, the test accuracy of a GBT trained on the removed features (i.e the features below a certain feature importance threshold) is very low up to 20\% of features removed, and quite low until 50\%, which suggests that most of these features are uninformative, and not solely redundant.

\paragraph{MLP-like architectures are not robust to uninformative features} In the two experiments shown in Fig. \ref{fig:features}, we can see that \textit{removing} uninformative features (\ref{fig:features}a) reduces the performance gap between MLPs (Resnet) and the other models (FT Transformers and tree-based models), while \textit{adding} uninformative features widens the gap. This shows that MLPs are less robust to uninformative features, and, given the frequency of such features in tabular datasets, partly explain the results from Sec. \ref{results}.

In Fig. \ref{fig:features}a, we also remove informative features as we remove a larger fraction of features. Our reasoning, which is backed by \ref{fig:features}b, is that the decrease in accuracy due to the removal of these features is compensated by the removal of uninformative features, which is more helpful for MLPs than for other models (we also remove redundant features at the same time, which should not impact our models)

\subsection{Finding 3: Data are non invariant by rotation, so should be learning procedures}


Why are MLPs much more hindered by uninformative features, compared to other models? One answer is that this learner is rotationally invariant in the sense of \cite{ngFeatureSelectionVs2004}: the learning procedure which learns an MLP on a training set and evaluate it on a testing set is unchanged when applying a rotation (unitary matrix) to the features on both the training and testing set. Indeed, \cite{ngFeatureSelectionVs2004} shows that any rotationallly invariant learning procedure has a worst-case sample complexity that grows at least linearly in the number of irrelevant features. Intuitively, to remove uninformative features, a rotationaly invariant algorithm has to first find the original orientation of the features, and then select the least informative ones: the information contained in the orientation of the data is lost. 

\begin{figure}
\begin{minipage}{.5\textwidth}
    \includegraphics[width=\linewidth]{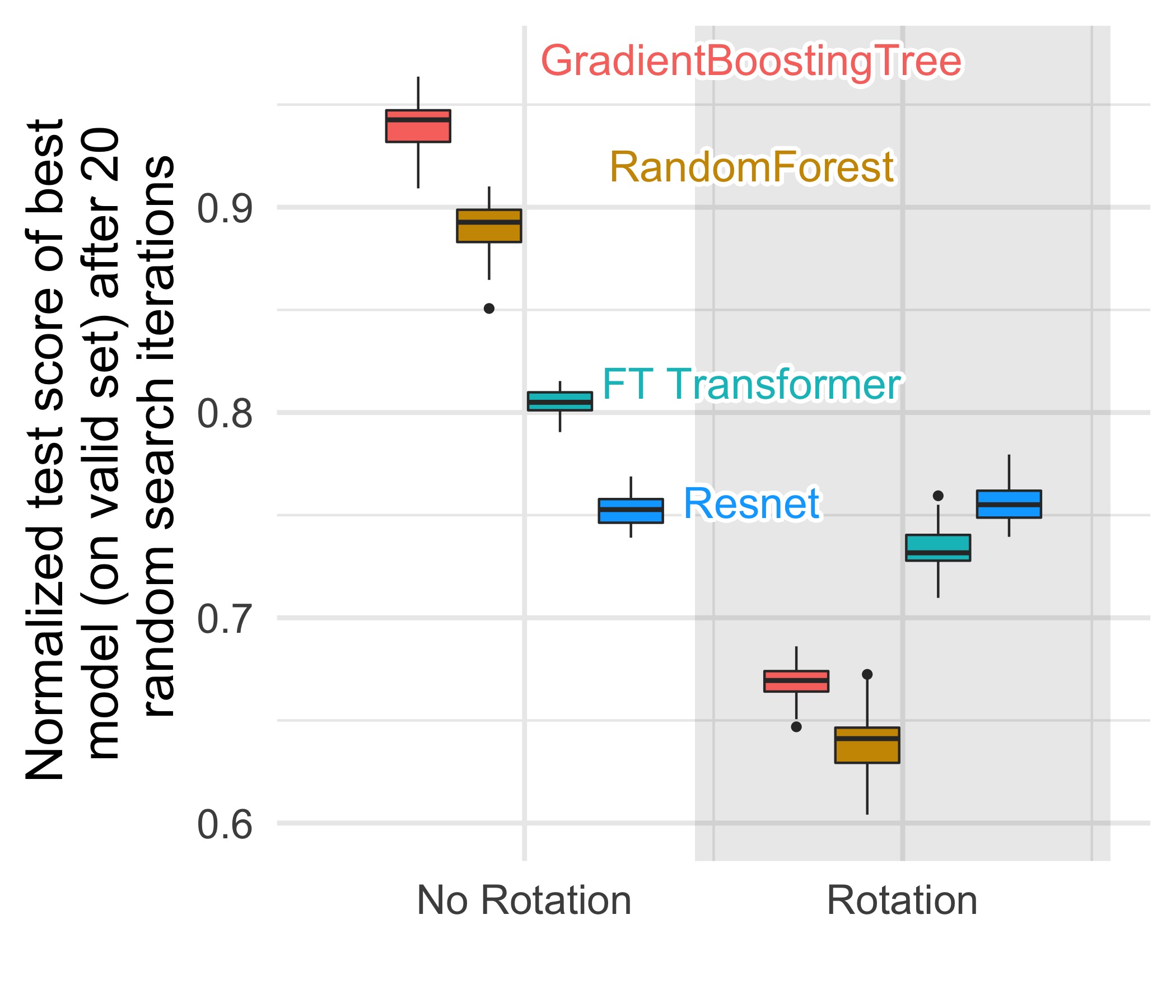}%
        \llap{\raisebox{.82\linewidth}{\parbox{\linewidth}{\sffamily\small{\textbf a.} With all features}}}%
    \label{fig:rotation_all_features}
\end{minipage}%
\begin{minipage}{.5\textwidth}
  \includegraphics[width=\linewidth]{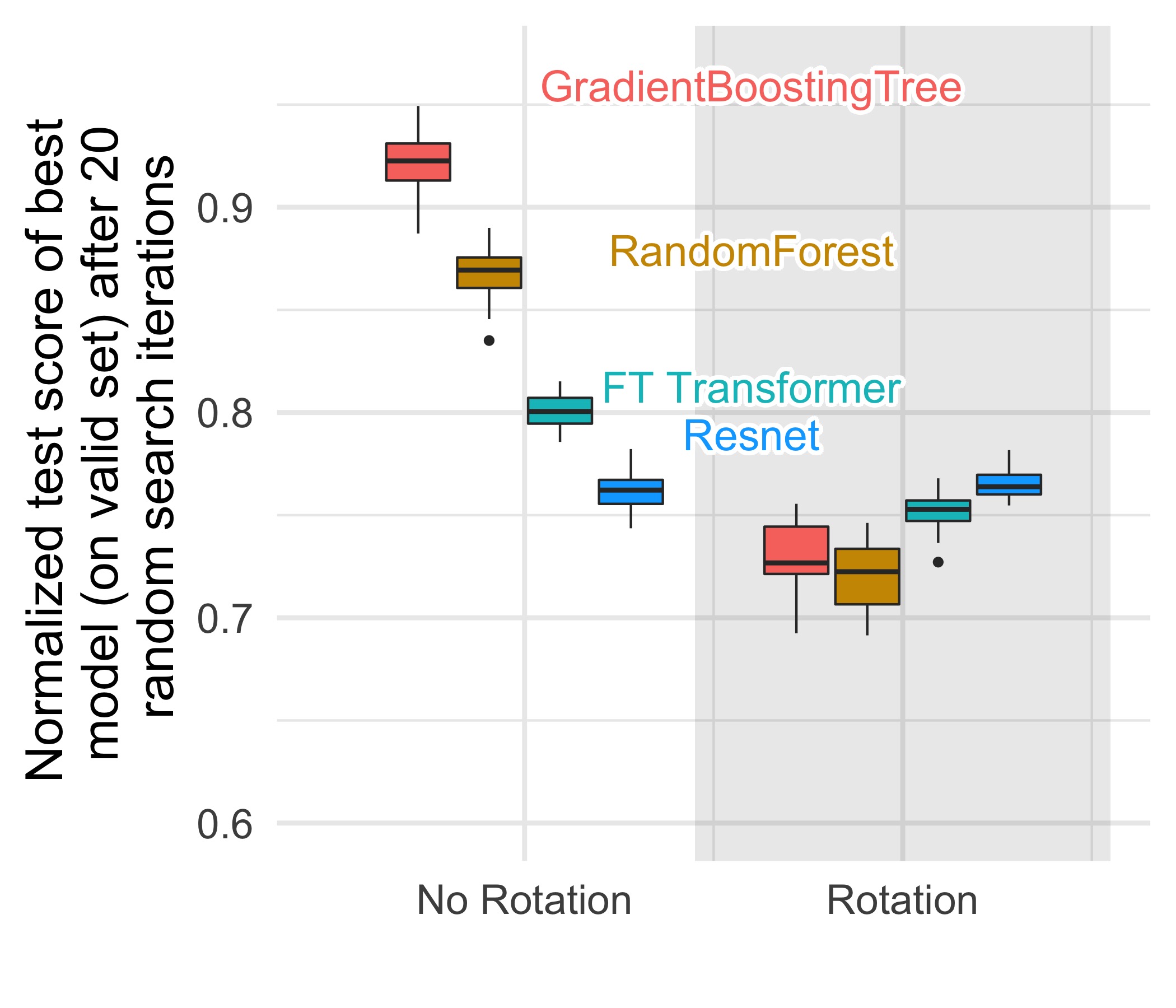}%
        \llap{\raisebox{.82\linewidth}{\parbox{\linewidth}{\sffamily\small{\textbf b.} With 50\% features removed}}}%
  \label{fig:rotation_features_removed}
\end{minipage}
\caption{\textbf{Normalized test accuracy of different models when randomly rotating our datasets}. Here, the classification benchmark on numerical features was used. All features are Gaussianized before the random rotations. The scores are averaged across datasets, and the boxes depict the distribution across random search shuffles. Right: the features are removed before data rotation.}%
\label{fig:rotation}%
\end{figure}

Fig. \ref{fig:rotation}a, which shows the change in test accuracy when randomly rotating our datasets, confirms that only Resnets are rotationally invariant. More striking, random rotations reverse the performance order: NNs are now above tree-based models and Resnets above FT Transformers. This suggests that rotation invariance is not desirable: similarly to vision~\citep{krizhevsky2012imagenet}, there is a natural basis (here, the original basis) which encodes best data-biases, and which can not be recovered by models invariant to rotations which potentially mixes features with very different statistical properties. Indeed, features of a tabular data typically carry meanings individually, as expressed by column names: {\tt age}, {\tt weight}. The link with uninformative features is apparent in \ref{fig:rotation}b: removing the least important half of the features in each dataset (before rotating), drops the performance of all models except Resnets, but the decrease is less significant than when using all features.


Our findings shed light on the results of \cite{somepalliSAINTImprovedNeural2021} and \cite{gorishniyEmbeddingsNumericalFeatures2022}, which add an embedding layer, even for numerical features, before MLP or Transformer models. Indeed, this layer breaks rotation invariance. The fact that very different types of embeddings seem to improve performance suggests that the sheer presence of an embedding which breaks the invariance is a key part of these improvements. We note that a promising avenue for further research would be to find other ways to break rotation invariance which might be less computationally costly than embeddings.

\section{Discussion and conclusion}\label{limitation}

\paragraph{Limitation}
Our study leaves open questions for future work: which other inductive biases of tree-based models explain their performances on tabular data? How would our evaluation change on very small datasets?  On very large datasets? What is the best way to handle specific challenges like missing data or high-cardinality categorical features, for NNs and tree-based models? With these best methods, how would the evaluation change including missing data?

\paragraph{Conclusion} While each publication on learning architectures for tabular data comes to different results using a different benchmarking methodology, our systematic benchmark, going beyond the specificities of a handful of datasets and accounting for hyper-parameter choice, reveals clear trends. On such data, tree-based models more easily yield good predictions, with much less computational cost. This superiority is explained by specific features of tabular data: irregular patterns in the target function, uninformative features, and non rotationally-invariant data where linear combinations of features misrepresent the information. Beyond these conclusions, our benchmark is reusable, allowing researchers to use our methodology and datasets for new architectures, and to easily compare them to those we explored via the shared benchmark raw results.
We hope that this benchmark will stimulate tabular deep-learning research and foster more thorough empirical evaluation of contributions.

\begin{ack}
GV and LG acknowledge support in part by the French Agence Nationale de la Recherche under Grant ANR-20-CHIA-0026 (LearnI). EO was supported by the Project ANR-21-CE23-0030 ADONIS and EMERG-ADONIS from Alliance SU.
\end{ack}


\bibliographystyle{unsrtnat}
\bibliography{neurips2022.bib} 

\newpage

\appendix


\section{Appendix}

\subsection{Datasets used}\label{supp:datasets}

We describe below all datasets used in our benchmarks, along with the link to the original dataset, as well as a new OpenML link to the transformed datasets used for our benchmarks. All datasets considered for the benchmarks, as well as the reason for their selection or their exclusion, are available at this link: \url{https://docs.google.com/spreadsheets/d/1Mgh27upycFcd3B6uA7YJyB9Pd4Y3UuY1gBfI02EZUGM/edit?usp=sharing}. Instructions on how to use these datasets to benchmark your own algorithms are available at \ref{supp:howto}.

\subsubsection{Numerical classification}

OpenML benchmark: \url{https://www.openml.org/search?type=benchmark&study_type=task&sort=tasks_included&id=298}

{\tiny%
\rowcolors{2}{white}{gray!25}
\begin{filecontents*}{datasets_csv/datasets_numerical_classif.csv}
dataset_name;n_samples;n_features;original link;new link
electricity;38474;7;https://openml.org/d/151;https://www.openml.org/d/44120
covertype;566602;10;https://openml.org/d/293;https://www.openml.org/d/44121
pol;10082;26;https://openml.org/d/722;https://www.openml.org/d/44122
house_16H;13488;16;https://openml.org/d/821;https://www.openml.org/d/44123
kdd_ipums_la_97-small;5188;20;https://openml.org/d/993;https://www.openml.org/d/44124
MagicTelescope;13376;10;https://openml.org/d/1120;https://www.openml.org/d/44125
bank-marketing;10578;7;https://openml.org/d/1461;https://www.openml.org/d/44126
phoneme;3172;5;https://openml.org/d/1489;https://www.openml.org/d/44127
MiniBooNE;72998;50;https://openml.org/d/41150;https://www.openml.org/d/44128
Higgs;940160;24;https://openml.org/d/42769;https://www.openml.org/d/44129
eye_movements;7608;20;https://openml.org/d/1044;https://www.openml.org/d/44130
jannis;57580;54;https://openml.org/d/41168;https://www.openml.org/d/44131
credit;16714;10;https://www.kaggle.com/c/GiveMeSomeCredit/data?select=cs-training.csv;https://www.openml.org/d/44089
california;20634;8;https://www.dcc.fc.up.pt/~ltorgo/Regression/cal_housing.html;https://www.openml.org/d/44090
wine;2554;11;https://archive.ics.uci.edu/ml/datasets/wine+quality;https://www.openml.org/d/44091
\end{filecontents*}
\csvautobooktabular[respect underscore=true,separator=semicolon]{datasets_csv/datasets_numerical_classif.csv}%
}

Note that we noticed a bit late that the number of samples in the transformed \textit{wine} dataset was just below our threshold, and decided to keep it. 

\subsubsection{Numerical regression}

OpenML benchmark: \url{https://www.openml.org/search?type=benchmark&study_type=task&sort=tasks_included&id=297}

{\tiny%
\rowcolors{2}{white}{gray!25}
\begin{filecontents*}{datasets_csv/datasets_numerical_regression.csv}
dataset_name;n_samples;n_features;original link;new link
cpu_act;8192;21;https://openml.org/d/197;https://www.openml.org/d/44132
pol;15000;26;https://openml.org/d/201;https://www.openml.org/d/44133
elevators;16599;16;https://openml.org/d/216;https://www.openml.org/d/44134
isolet;7797;613;https://openml.org/d/300;https://www.openml.org/d/44135
wine_quality;6497;11;https://openml.org/d/287;https://www.openml.org/d/44136
Ailerons;13750;33;https://openml.org/d/296;https://www.openml.org/d/44137
houses;20640;8;https://openml.org/d/537;https://www.openml.org/d/44138
house_16H;22784;16;https://openml.org/d/574;https://www.openml.org/d/44139
diamonds;53940;6;https://openml.org/d/42225;https://www.openml.org/d/44140
Brazilian_houses;10692;8;https://openml.org/d/42688;https://www.openml.org/d/44141
Bike_Sharing_Demand;17379;6;https://openml.org/d/42712;https://www.openml.org/d/44142
nyc-taxi-green-dec-2016;581835;9;https://openml.org/d/42729;https://www.openml.org/d/44143
house_sales;21613;15;https://openml.org/d/42731;https://www.openml.org/d/44144
sulfur;10081;6;https://openml.org/d/23515;https://www.openml.org/d/44145
medical_charges;163065;5;https://openml.org/d/42720;https://www.openml.org/d/44146
MiamiHousing2016;13932;14;https://openml.org/d/43093;https://www.openml.org/d/44147
superconduct;21263;79;https://openml.org/d/43174;https://www.openml.org/d/44148
california;20640;8;https://www.dcc.fc.up.pt/~ltorgo/Regression/cal_housing.html;https://www.openml.org/d/44025
fifa;18063;5;https://www.kaggle.com/datasets/stefanoleone992/fifa-22-complete-player-dataset;https://www.openml.org/d/44026
year;515345;90;https://archive.ics.uci.edu/ml/datasets/yearpredictionmsd;https://www.openml.org/d/44027
\end{filecontents*}
\csvautobooktabular[respect underscore=true,separator=semicolon]{datasets_csv/datasets_numerical_regression.csv}%
}

\subsubsection{Categorical classification}

OpenML benchmark: \url{https://www.openml.org/search?type=benchmark&sort=date&study_type=task&id=300}

{\tiny%
\rowcolors{2}{white}{gray!25}
\begin{filecontents*}{datasets_csv/datasets_categorical_classif.csv}
dataset_name;n_samples;n_features;Original link;New link
electricity;38474;8;https://openml.org/d/151;https://www.openml.org/d/44156
eye_movements;7608;23;https://openml.org/d/1044;https://www.openml.org/d/44157
KDDCup09_upselling;5032;45;https://openml.org/d/1114;https://www.openml.org/d/44158
covertype;423680;54;https://openml.org/d/1596;https://www.openml.org/d/44159
rl;4970;12;https://openml.org/d/41160;https://www.openml.org/d/44160
road-safety;111762;32;https://openml.org/d/42803;https://www.openml.org/d/44161
compass ;16644;17;https://www.kaggle.com/datasets/danofer/compass?select=cox-violent-parsed.csv;https://www.openml.org/d/44162
\end{filecontents*}
\csvautobooktabular[respect underscore=true,separator=semicolon]{datasets_csv/datasets_categorical_classif.csv}%
}

\subsubsection{Categorical regression}

OpenML benchmark: \url{https://www.openml.org/search?type=benchmark&study_type=task&sort=tasks_included&id=299}

{\tiny%
\rowcolors{2}{white}{gray!25}
\begin{filecontents*}{datasets_csv/datasets_categorical_regression.csv}
dataset_name;n_features;n_samples;Original link;New link
yprop_4_1;62;8885;https://openml.org/d/416;https://www.openml.org/d/44054
analcatdata_supreme;7;4052;https://openml.org/d/504;https://www.openml.org/d/44055
visualizing_soil;4;8641;https://openml.org/d/688;https://www.openml.org/d/44056
black_friday;9;166821;https://openml.org/d/41540;https://www.openml.org/d/44057
diamonds;9;53940;https://openml.org/d/42225;https://www.openml.org/d/44059
Mercedes_Benz_Greener_Manufacturing;359;4209;https://openml.org/d/42570;https://www.openml.org/d/44061
Brazilian_houses;11;10692;https://openml.org/d/42688;https://www.openml.org/d/44062
Bike_Sharing_Demand;11;17379;https://openml.org/d/42712;https://www.openml.org/d/44063
OnlineNewsPopularity;59;39644;https://openml.org/d/42724;https://www.openml.org/d/44064
nyc-taxi-green-dec-2016;16;581835;https://openml.org/d/42729;https://www.openml.org/d/44065
house_sales;17;21613;https://openml.org/d/42731;https://www.openml.org/d/44066
particulate-matter-ukair-2017;6;394299;https://openml.org/d/42207;https://www.openml.org/d/44068
SGEMM_GPU_kernel_performance;9;241600;https://openml.org/d/43144;https://www.openml.org/d/44069
\end{filecontents*}
\csvautobooktabular[respect underscore=true,separator=semicolon]{datasets_csv/datasets_categorical_regression.csv}%
}

\subsection{More benchmarks}\label{supp:benchmarks}

\subsubsection{Results as a function of random search time}

\begin{figure}
\begin{minipage}{.5\textwidth}
  \includegraphics[width=\linewidth]{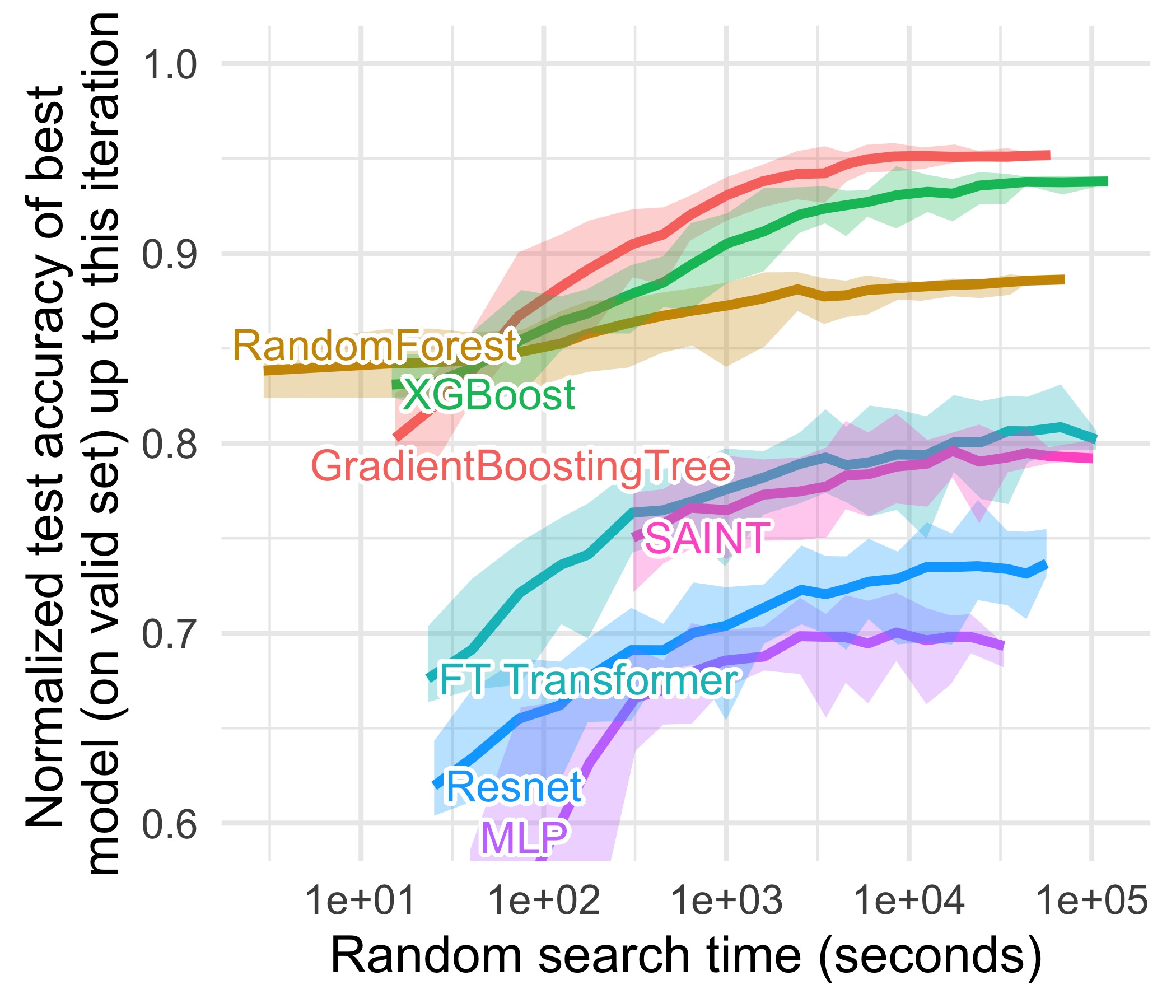}%
  \llap{\raisebox{.8\linewidth}{\parbox{.8\linewidth}{\sffamily{\bfseries Classification} (15 datasets)}}}%
\end{minipage}%
\begin{minipage}{.5\textwidth}
  \includegraphics[width=\linewidth]{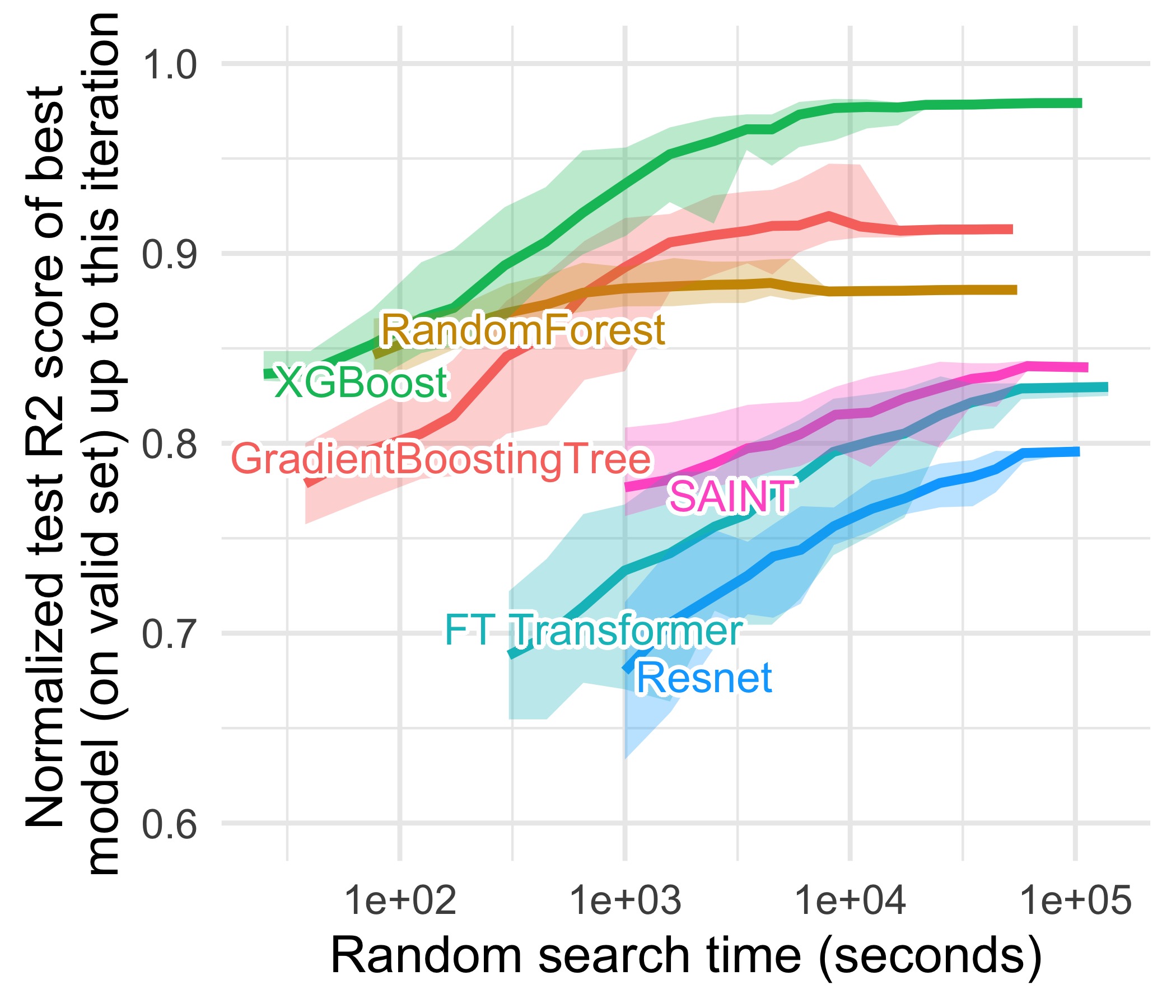}%
    \llap{\raisebox{.8\linewidth}{\parbox{.8\linewidth}{\sffamily{\bfseries Regression} (18 datasets)}}}%
\end{minipage}
\caption{\textbf{Time benchmark on medium-sized datasets, with only numerical features}. The first random search iteration corresponds to default hyperparameters. Each value corresponds to the test score of the best model (on the validation set) after a specific time spent doing random search, averaged on 15 shuffles of the random search order. The ribbon corresponds to the minimum and maximum scores on these 15 shuffles.}
\label{fig:benchmark_numeric_time}

\bigskip

\begin{minipage}{.5\textwidth}
  \includegraphics[width=\linewidth]{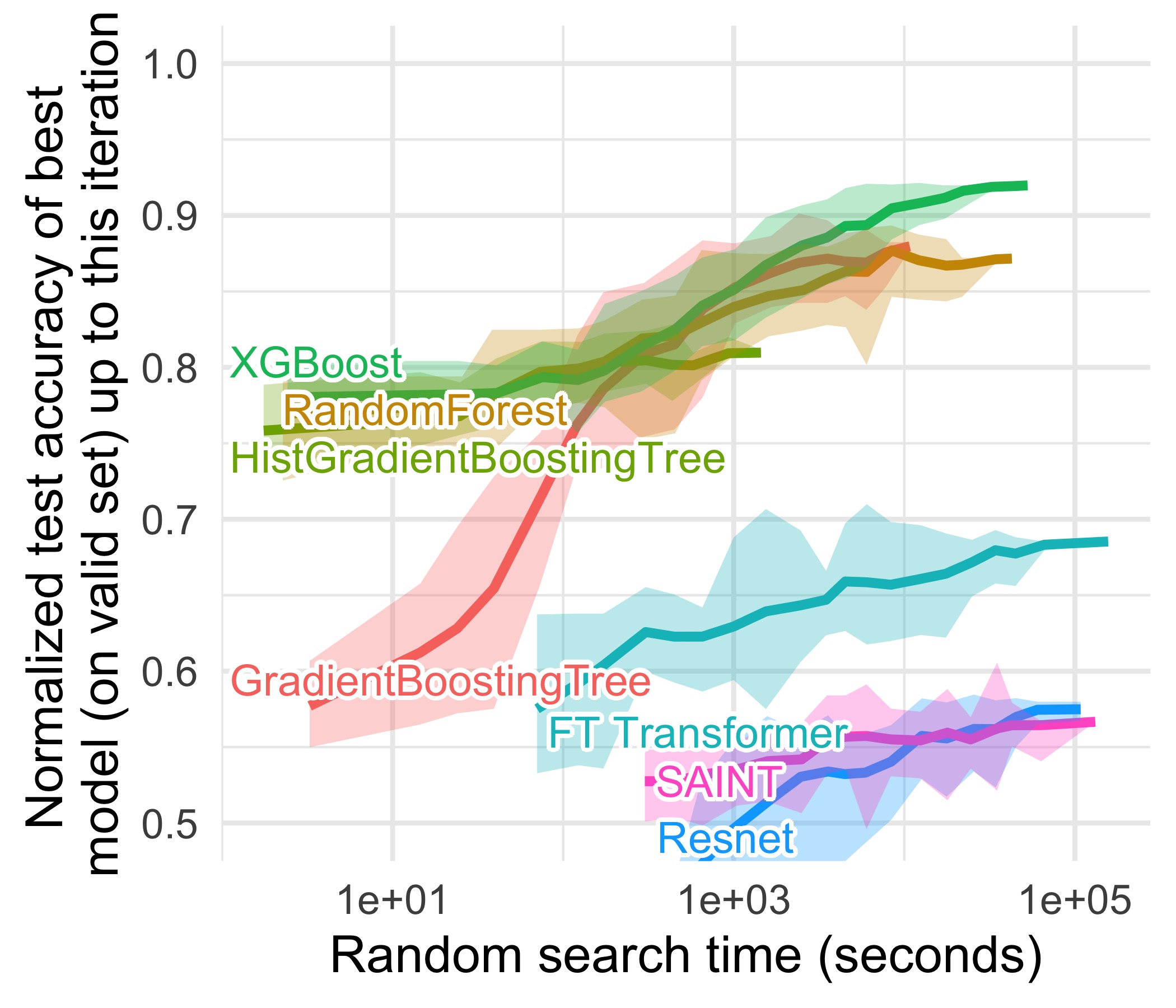}%
    \llap{\raisebox{.82\linewidth}{\parbox{.8\linewidth}{\sffamily{\bfseries Classification} (7 datasets)}}}%
\end{minipage}%
\begin{minipage}{.5\textwidth}
  \includegraphics[width=\linewidth]{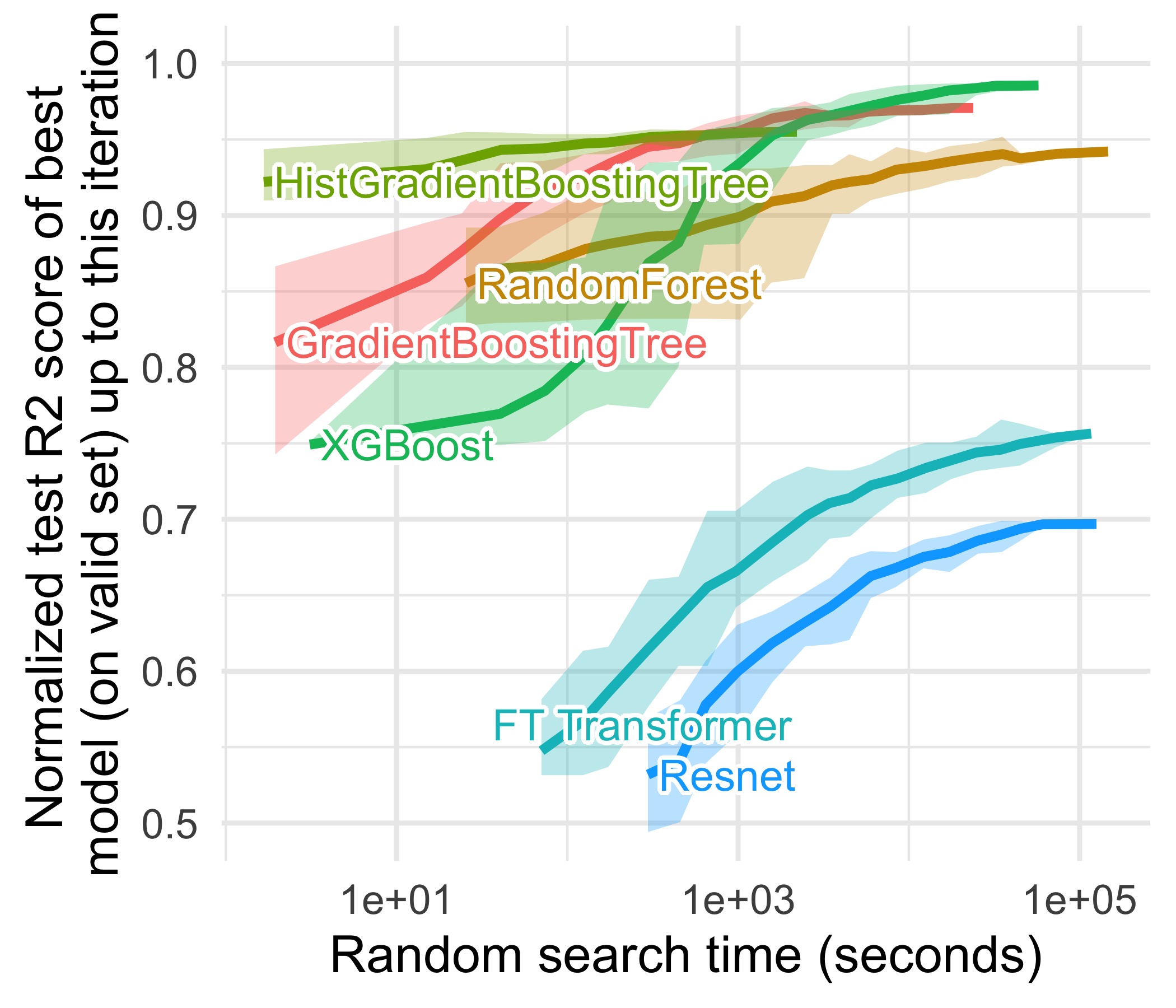}%
      \llap{\raisebox{.82\linewidth}{\parbox{.8\linewidth}{\sffamily{\bfseries Regression} (14 datasets)}}}%
\end{minipage}
\caption{\textbf{Time benchmark on medium-sized datasets, with both numerical and categorical features}. The first random search iteration corresponds to default hyperparameters. Each value corresponds to the test score of the best model (on the validation set) after a specific time spent doing random search, averaged on 15 shuffles of the random search order. The ribbon corresponds to the minimum and maximum scores on these 15 shuffles.}
\label{fig:benchmark_categorical_time}
\end{figure}

In Figure \ref{fig:benchmark_numeric_time} and Figure \ref{fig:benchmark_categorical_time}, we present the same results that in section \ref{results}, but as a function of random search \textit{time} instead of random search \textit{iterations}.

\paragraph{Details} Evaluation and training time are added. Time is averaged among folds, and cumulative time spent on random search is binned into 20 bins. Deep learning models are run on GPUs, and tree-based models on CPUs (see \ref{supp:benchmark_details}). We present this comparison to give a rough sense of the speed difference between tree-based models and neural networks, but this should not be considered a rigorous comparison of the speed of different models, as we use different types of GPUs and CPUs.

\paragraph{Results} Looking at the results as a function of random search time rather than random search iterations makes tree-based models superiority even more striking. Neural networks and tree-based models were close for some benchmarks after a small number of iterations, but for the same amount of time spent on random search, tree-based models scores are always high above neural networks.

\clearpage

\subsubsection{Large-sized datasets}

\begin{figure}
\begin{minipage}{.5\textwidth}
  \includegraphics[width=\linewidth]{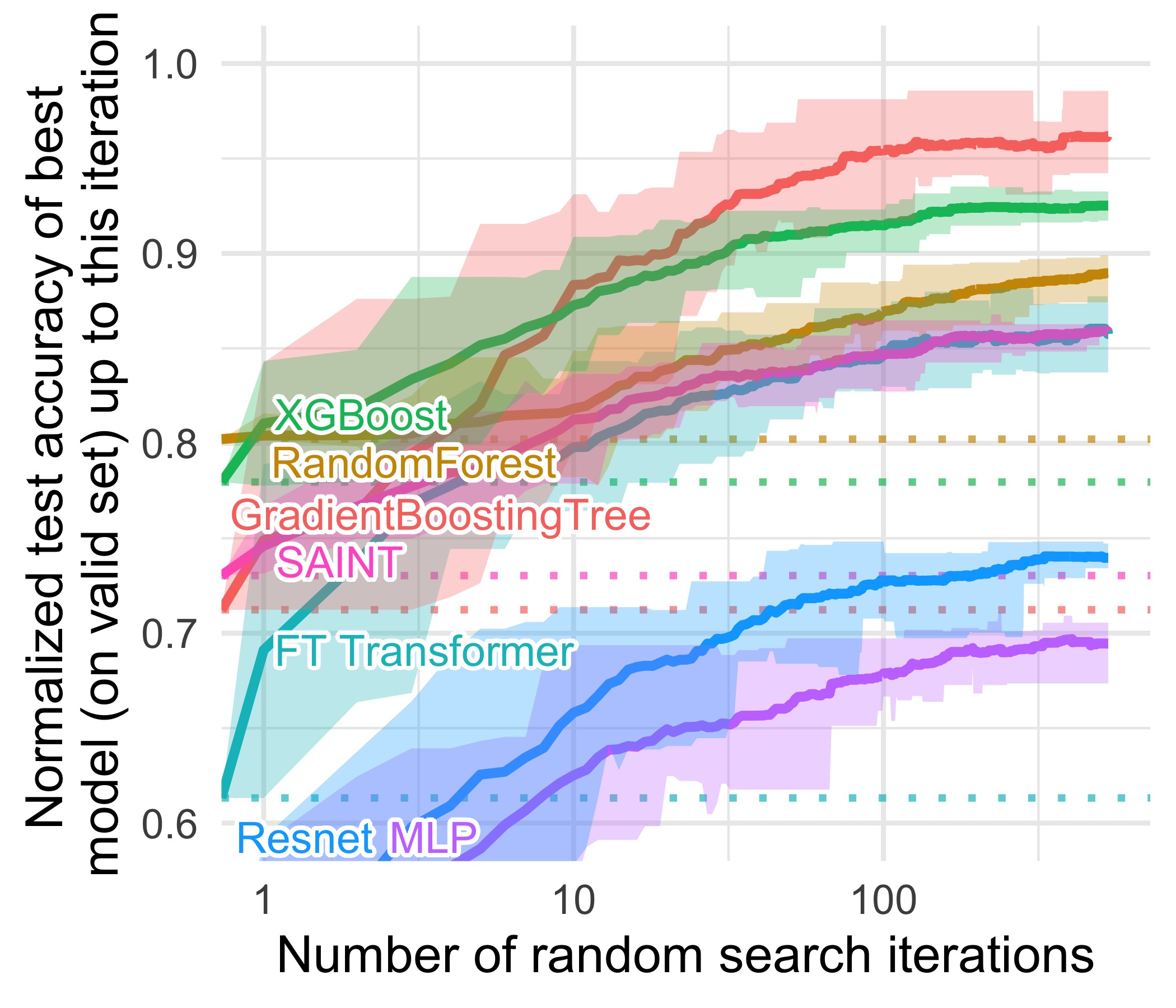}%
    \llap{\raisebox{.8\linewidth}{\parbox{.8\linewidth}{\sffamily{\bfseries Medium} sized train set}}}%
\end{minipage}%
\begin{minipage}{.5\textwidth}
  \includegraphics[width=\linewidth]{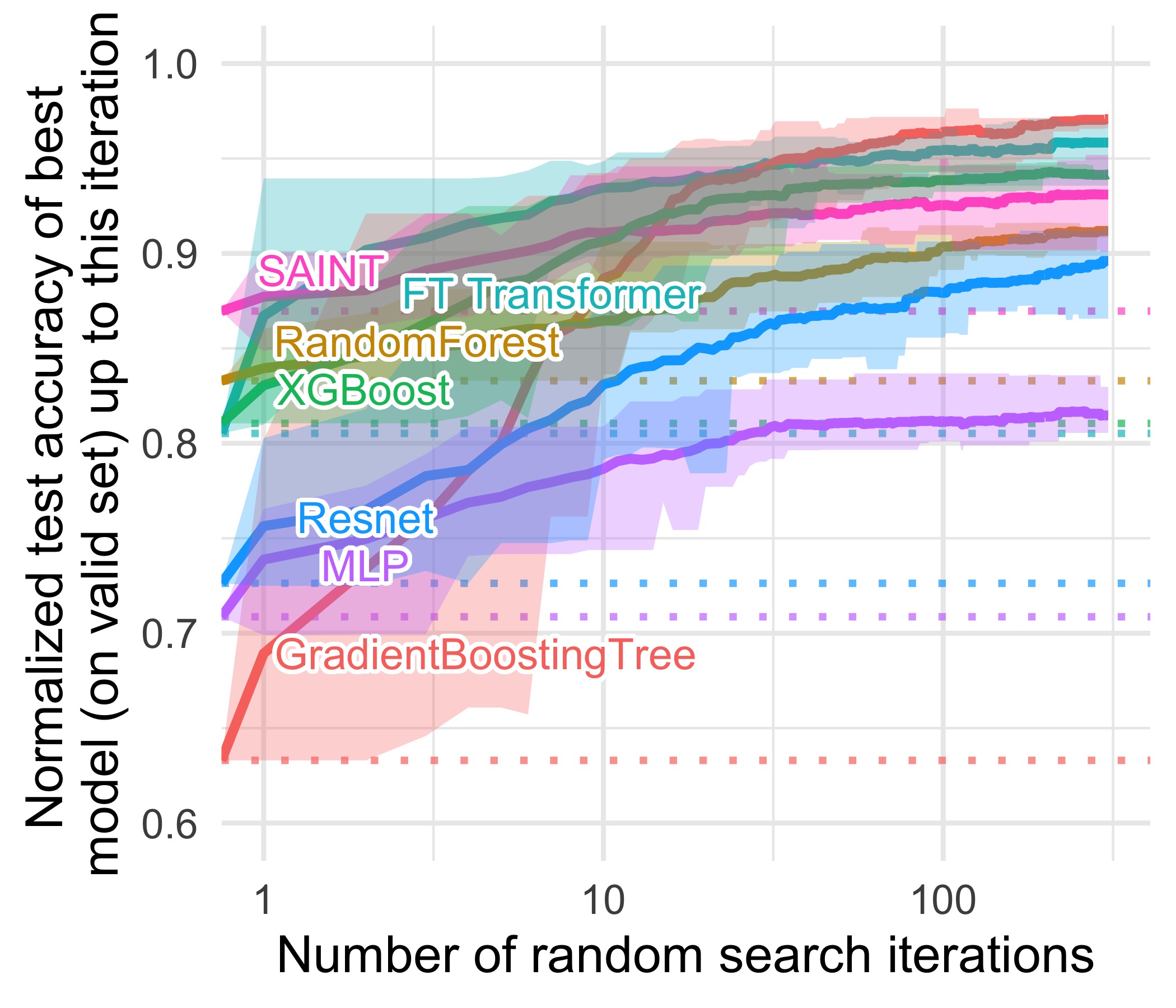}%
      \llap{\raisebox{.8\linewidth}{\parbox{.8\linewidth}{\sffamily{\bfseries Large} sized train set}}}%
\end{minipage}
\caption{\textbf{Comparison of accuracies on 4 classification tasks for different train set sizes, with only numerical features}. Only datasets with more than 50,000 samples were kept, and the train set size was truncated to either 10,000 samples or 50,000 samples. Dotted lines correspond to the score of the default hyperparameters, which is also the first random search iteration. Each value corresponds to the test score of the best model (on the validation set) after a specific number of random search iterations, averaged on 15 shuffles of the random search order. The ribbon corresponds to the minimum and maximum scores on these 15 shuffles.}
\label{fig:benchmark_numerical_classif_large}

\bigskip

\begin{subfigure}{.5\textwidth}
  \includegraphics[width=\linewidth]{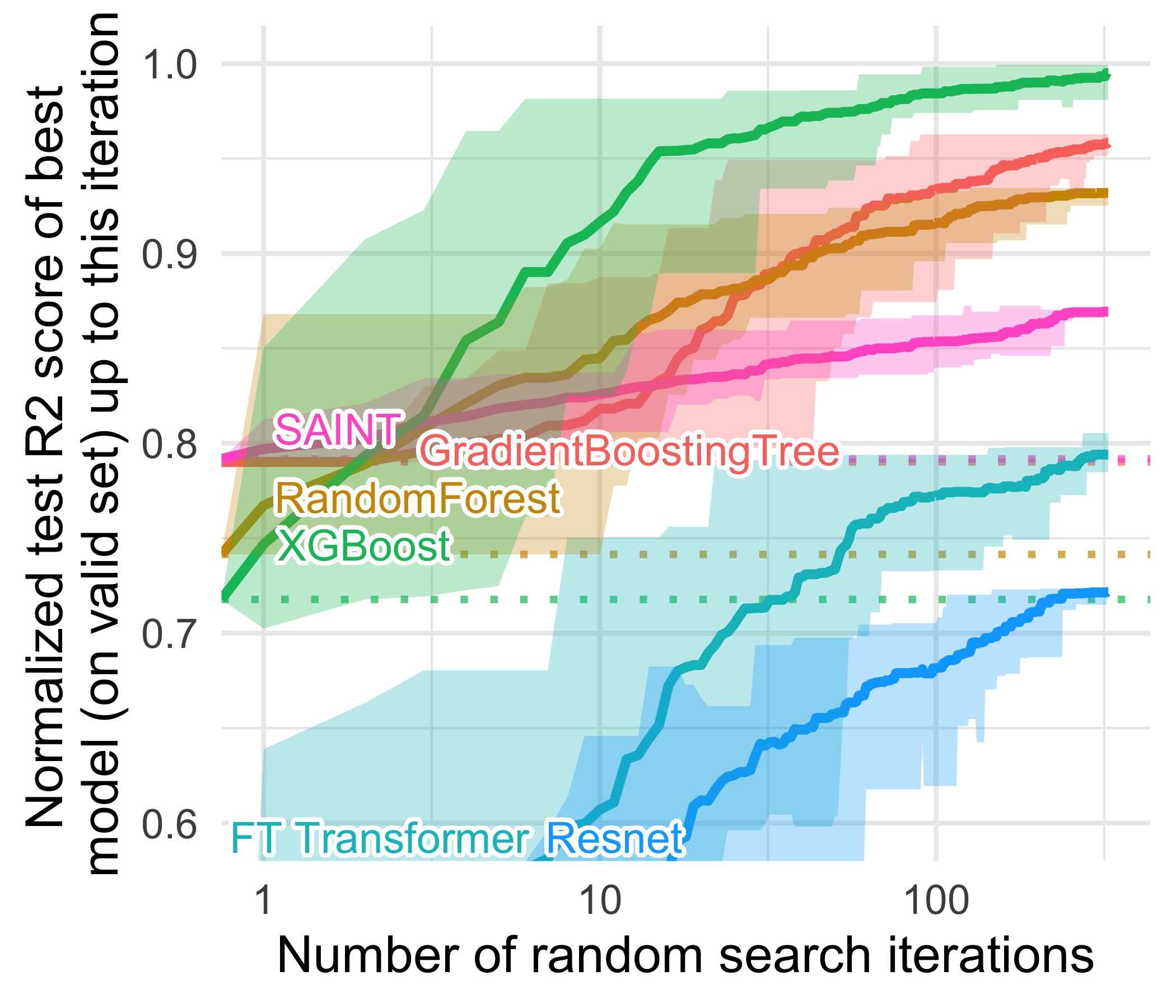}%
      \llap{\raisebox{.8\linewidth}{\parbox{.8\linewidth}{\sffamily{\bfseries Medium} sized train set}}}%
\end{subfigure}%
\begin{subfigure}{.5\textwidth}
  \includegraphics[width=\linewidth]{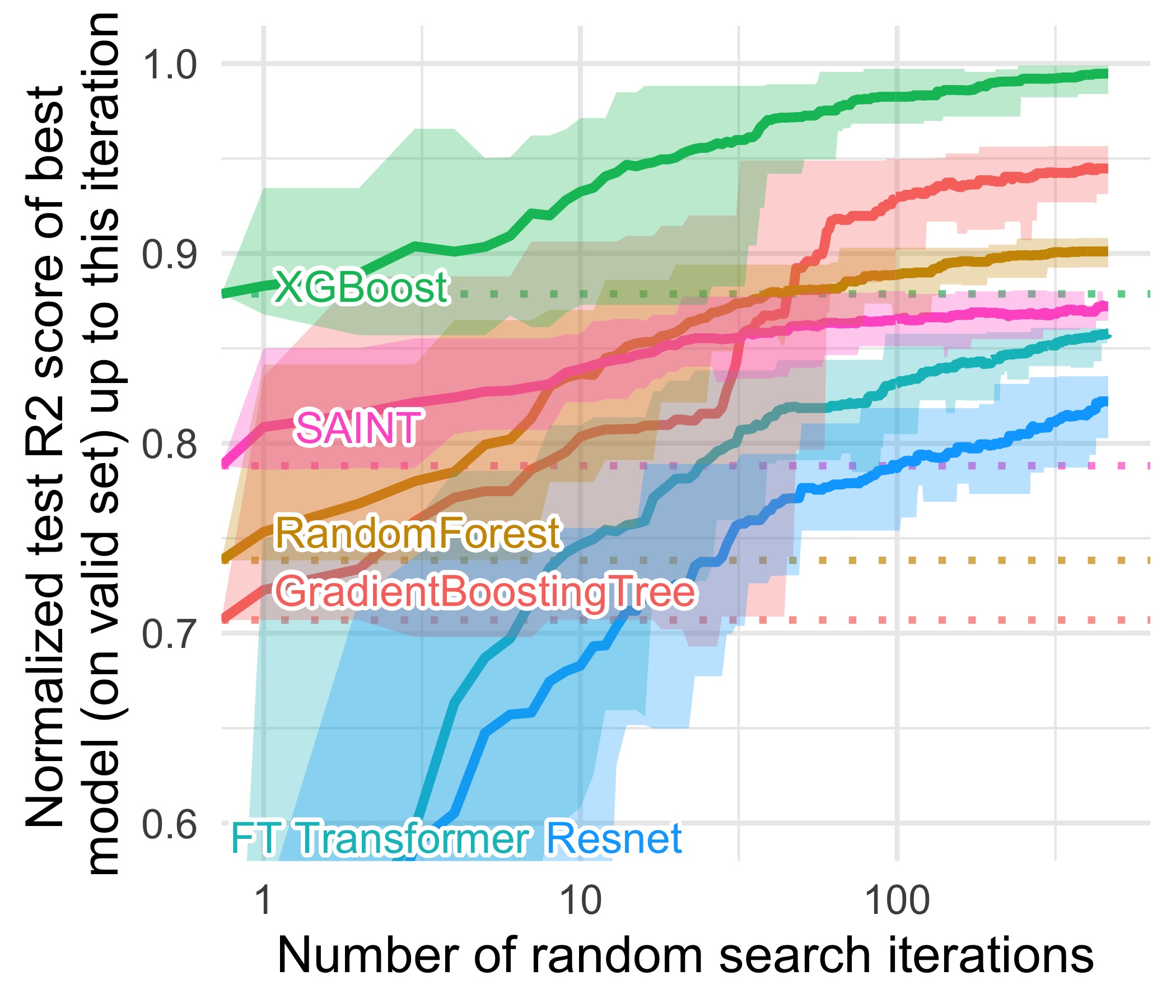}%
        \llap{\raisebox{.8\linewidth}{\parbox{.8\linewidth}{\sffamily{\bfseries Large} sized train set}}}%
\end{subfigure}
\caption{\textbf{Comparison of R2 scores on 3 regression tasks for different train set sizes, with only numerical features}. Only datasets with more than 50,000 samples were kept, and the train set size was truncated to either 10,000 samples or 50,000 samples. Dotted lines correspond to the score of the default hyperparameters, which is also the first random search iteration. Each value corresponds to the test score of the best model (on the validation set) after a specific number of random search iterations, averaged on 15 shuffles of the random search order. The ribbon corresponds to the minimum and maximum scores on these 15 shuffles.}
\label{fig:benchmark_numerical_regression_large}
\end{figure}

\begin{figure}
\begin{subfigure}{.5\textwidth}
  
  \includegraphics[width=\linewidth]{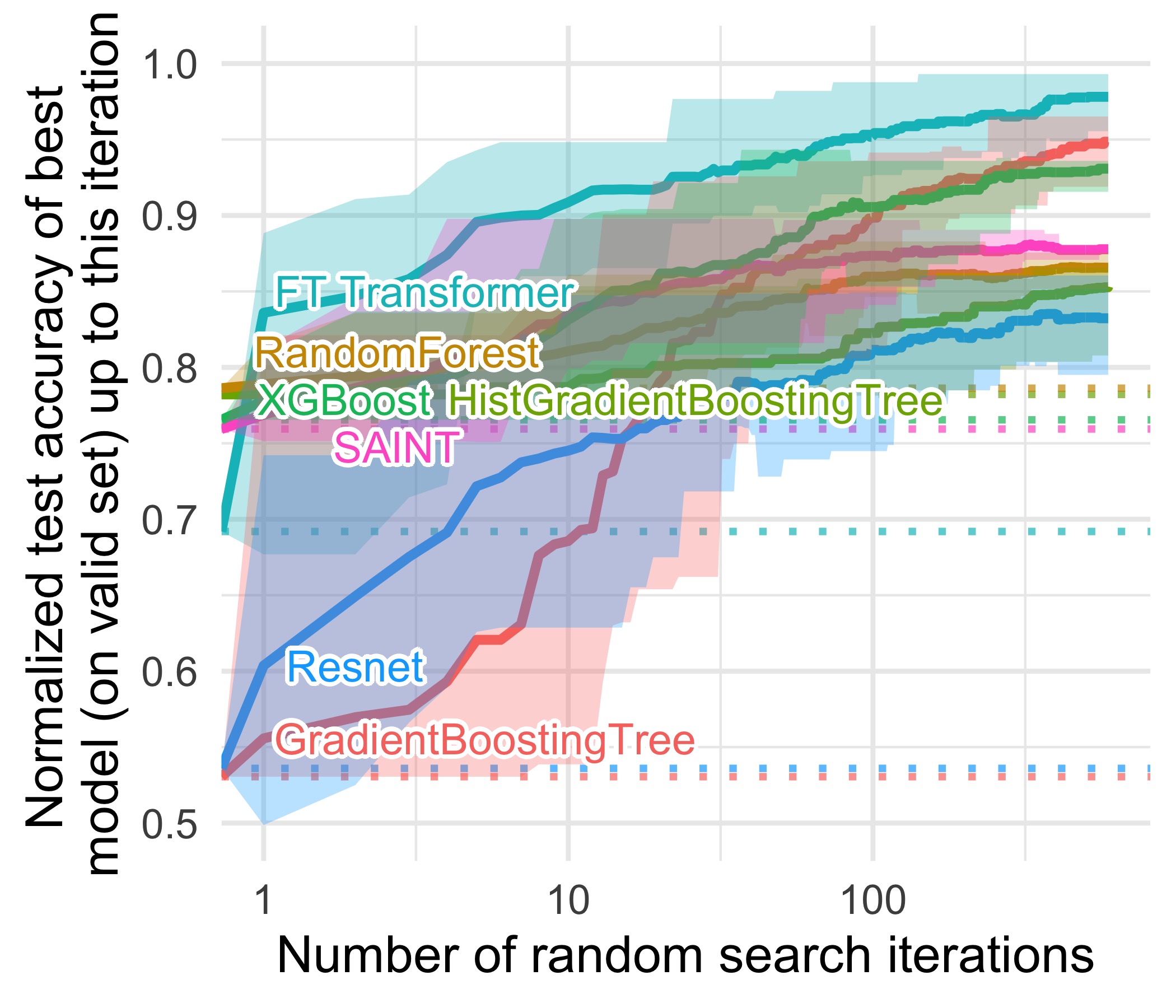}%
        \llap{\raisebox{.8\linewidth}{\parbox{.8\linewidth}{\sffamily{\bfseries Medium} sized train set}}}%
\end{subfigure}%
\begin{subfigure}{.5\textwidth}
  \includegraphics[width=\linewidth]{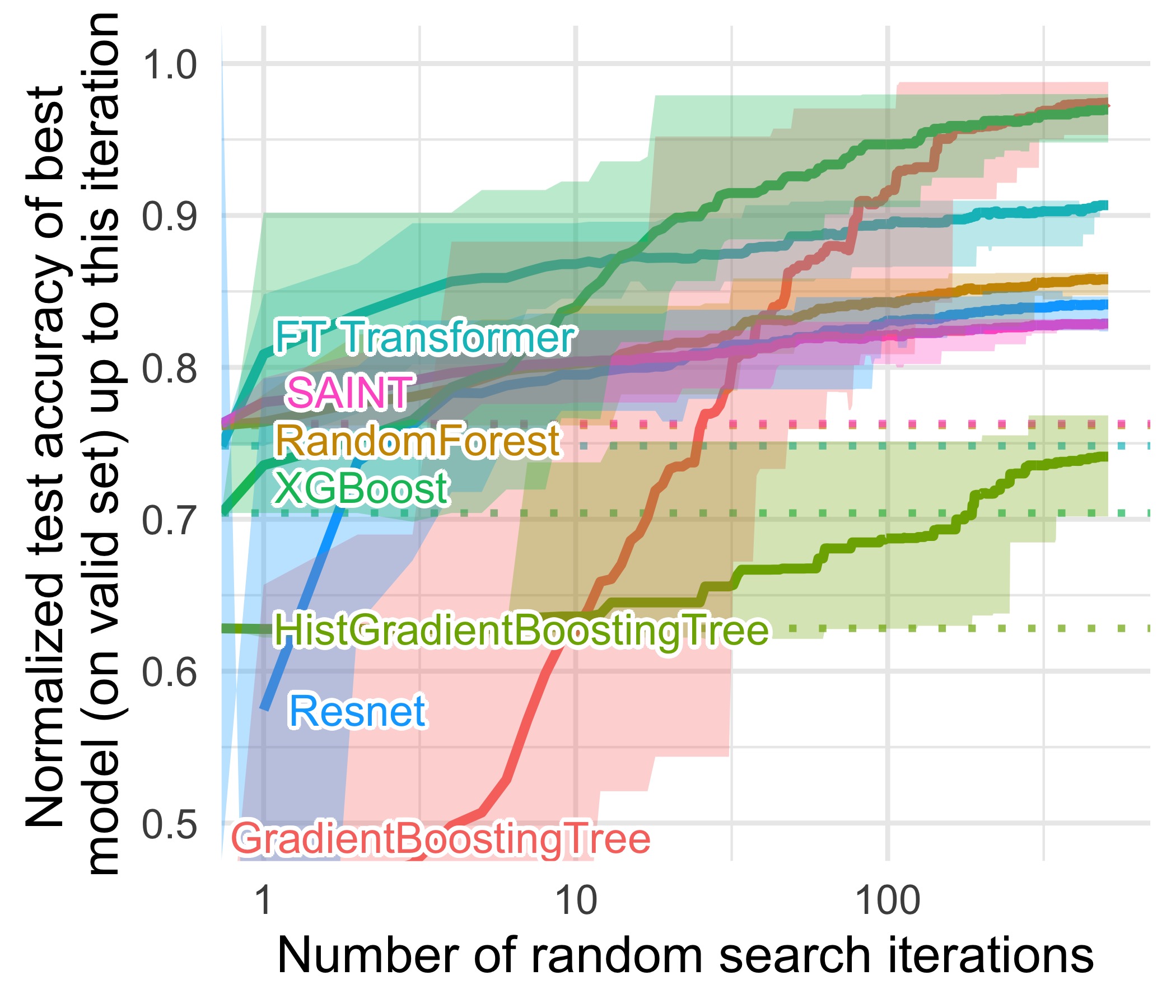}%
          \llap{\raisebox{.8\linewidth}{\parbox{.8\linewidth}{\sffamily{\bfseries Large} sized train set}}}%
\end{subfigure}
\caption{\textbf{Comparison of accuracies on 2 classification tasks for different train set sizes, with both numerical and categorical features}. Only datasets with more than 50,000 samples were kept, and the train set size was truncated to either 10,000 samples or 50,000 samples. Dotted lines correspond to the score of the default hyperparameters, which is also the first random search iteration. Each value corresponds to the test score of the best model (on the validation set) after a specific number of random search iterations, averaged on 15 shuffles of the random search order. The ribbon corresponds to the minimum and maximum scores on these 15 shuffles.}
\label{fig:benchmark_categorical_classif_large}

\bigskip

\begin{subfigure}{.5\textwidth}
  \includegraphics[width=\linewidth]{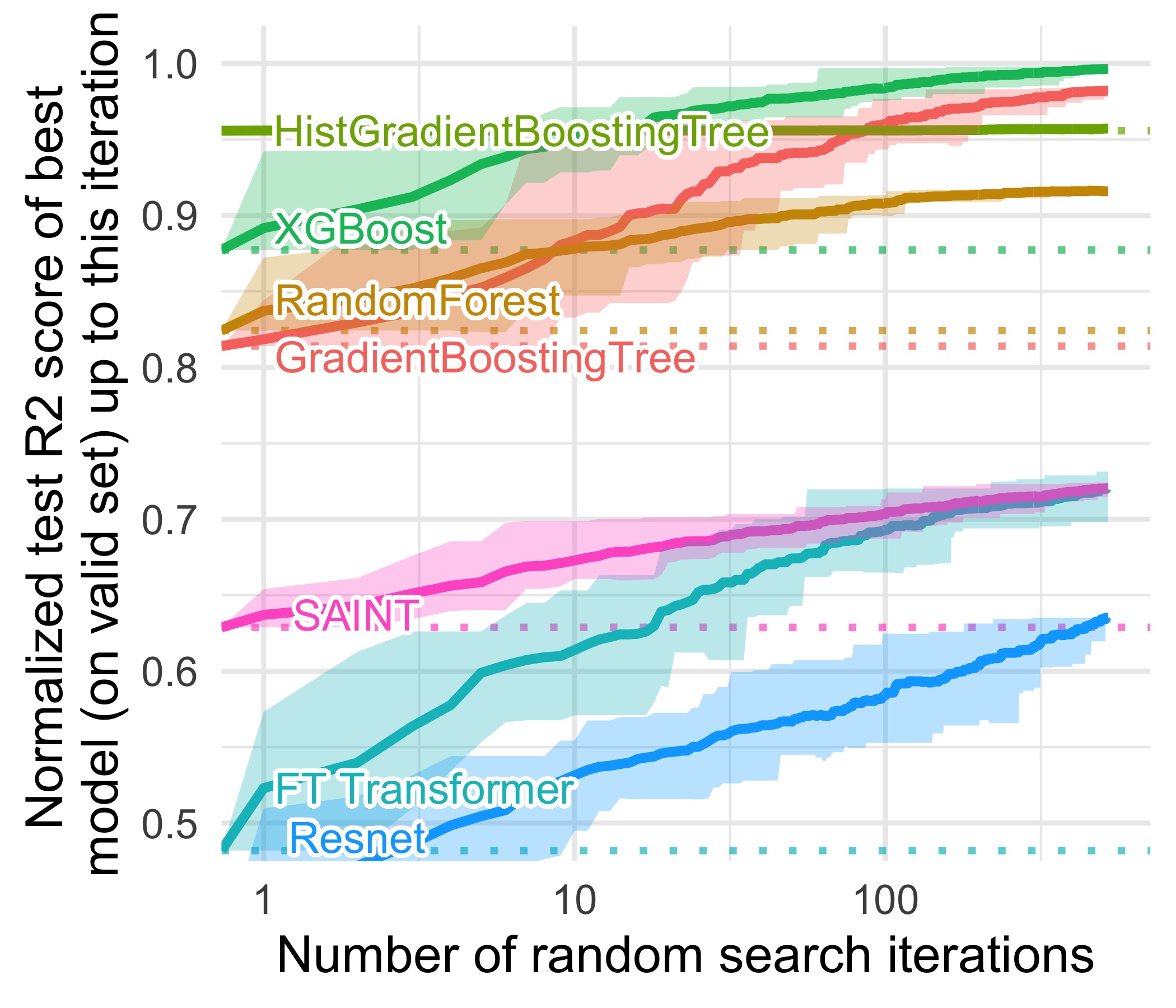}%
        \llap{\raisebox{.8\linewidth}{\parbox{.8\linewidth}{\sffamily{\bfseries Medium} sized train set}}}%
\end{subfigure}%
\begin{subfigure}{.5\textwidth}
  \includegraphics[width=\linewidth]{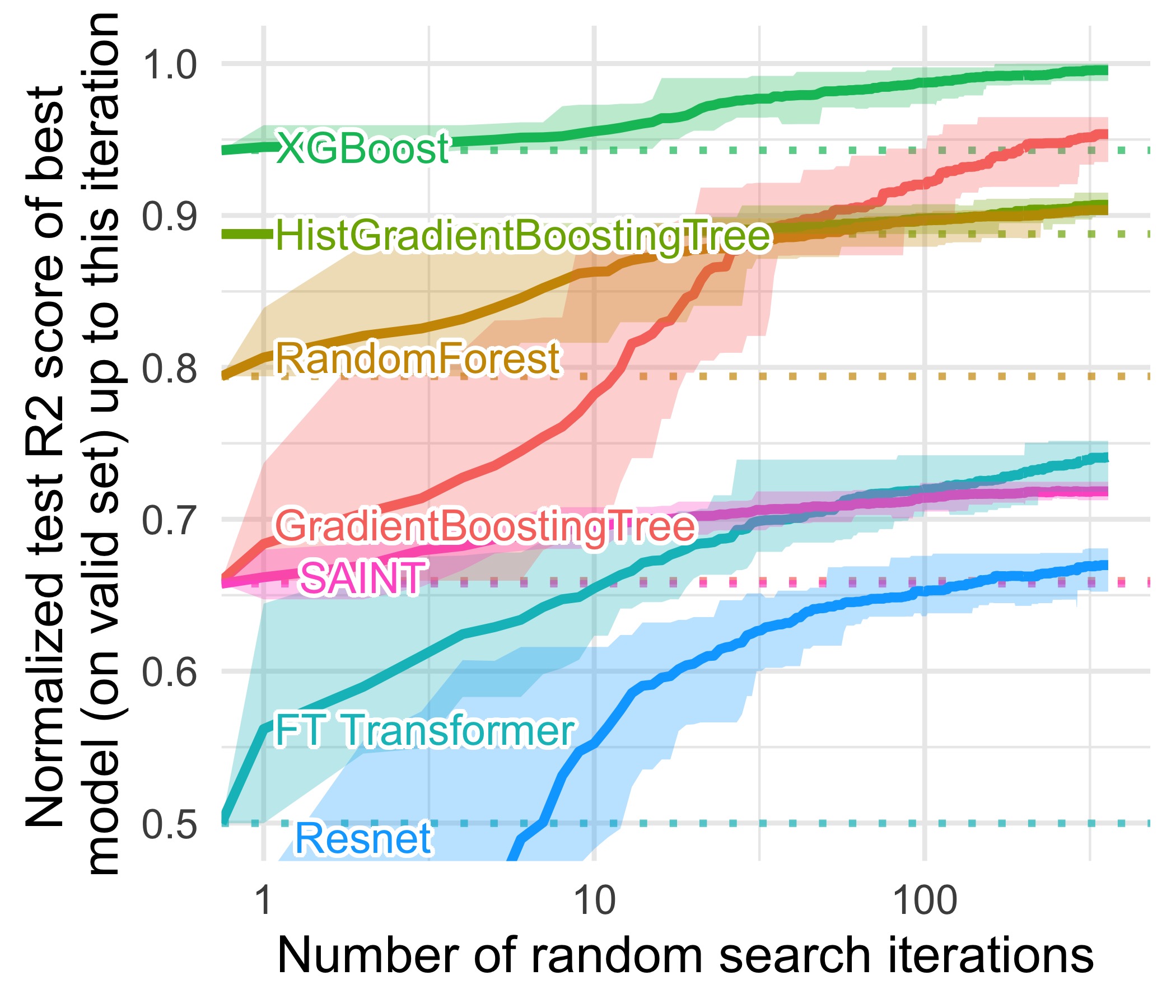}%
          \llap{\raisebox{.8\linewidth}{\parbox{.8\linewidth}{\sffamily{\bfseries Large} sized train set}}}%
\end{subfigure}
\caption{\textbf{Comparison of R2 scores on 5 regression tasks for different train set sizes, with both numerical and categorical features}. Only datasets with more than 50,000 samples were kept, and the train set size was truncated to either 10,000 samples or 50,000 samples. Dotted lines correspond to the score of the default hyperparameters, which is also the first random search iteration. Each value corresponds to the test score of the best model (on the validation set) after a specific number of random search iterations, averaged on 15 shuffles of the random search order. The ribbon corresponds to the minimum and maximum scores on these 15 shuffles.}
\label{fig:benchmark_categorical_regression_large}
\end{figure}

We extend our benchmark to large-scale datasets: in Figures \ref{fig:benchmark_numerical_classif_large}, \ref{fig:benchmark_numerical_regression_large}, \ref{fig:benchmark_categorical_classif_large} and \ref{fig:benchmark_categorical_regression_large}, we compare the results of our models on the same set of datasets, in large-size (train set truncated to 50,000 samples) and medium-size (train set truncated to 10,000 samples) settings.

We only keep datasets with more than 50,000 samples and restrict the train set size to 50,000 samples (vs 10,000 samples for the medium-sized benchmark). Unfortunately, this excludes a lot of datasets, which makes the comparison less clear. However, it seems that, in most cases, increasing the train set size reduces the gap between neural networks and tree-based models. We leave a rigorous study of this trend to future work.

\clearpage

\subsection{More details on benchmark}\label{supp:benchmark_details}

\paragraph{Train / Validation / Test split} We take 70\% of samples for the train set (or the percentage which corresponds to the maximum train set size if 70\% is too high). Of the remaining 30\%, we take 30\% for the validation set, and 70\% for the test set. The validation and test sets are truncated to 50,000 samples for speed. Note that the validation set is only used to select the best performing hyperparameter combination during the random search, and is distinct from the validation set used for early stopping (which is part of the train set).

\paragraph{Number of folds}

For each dataset and hyperparameters combination, we vary the number of folds used for our algorithms evaluation depending on the number of testing samples:

\begin{itemize}
 \item If We have more than 6000 samples, we evaluate our algorithms on 1 fold.
 \item If we have between 3000 and 6000 samples, we evaluate our algorithms on 2 folds.
 \item If we have between 1000 and 3000 samples, we evaluate our algorithms on 3 folds.
 \item If we have less than 1000 testing samples, we evaluate our algorithms on 5 folds.
\end{itemize}

Every algorithm and hyperparameters combination is evaluated on the same folds.

\paragraph{Hardware}

For all our benchmarks and experiments, we use the hardware below. The hardware was chosen based on availability, with Neural Networks always running on GPU and tree-based models running on CPU. 

GPUs: NVIDIA Quadro RTX 6000, NVIDIA TITAN Xp, NVIDIA A100, NVIDIA V100, NVIDIA Tesla T4, NVIDIA A40, NVIDIA TITAN RTX, NVIDIA TITAN V

CPUs: AMD EPYC 7742 64-Core Processor, AMD EPYC 7702 64-Core Processor, Intel(R) Xeon(R) CPU E5-2660 v2, Intel(R) Xeon(R) Gold 6226R CPU

\paragraph{Hyperparameters space}

Hyperparameters spaces are based on Hyperopt-sklearn \citep{komerHyperoptSklearnAutomaticHyperparameter2014} when available, from \cite{gorishniyRevisitingDeepLearning2021} and from \cite{borisovDeepNeuralNetworks2021}. We made some changes when combining sources, or when the original distribution was not compatible with Weight and Biases sweeps.

Default parameters for tree-based models are ScikitLearn's defaults. All neural networks are run for 300 epochs, with early stopping and checkpointing (the best model on the validation set is kept). Early stopping patience is 40 for MLP, Resnet, and FT Transformer, and 10 for SAINT.

\begin{table}[hbt!]
\rowcolors{2}{white}{gray!25}
\begin{tabular}{lll}
\hline Parameter & Distribution & Default \\
\hline Num layers & UniformInt $[1,6]$ & $3$ \\
Feature embedding size & UniformInt $[64,512]$ & $192$ \\
Residual dropout Uniform & $[0,0.5]$ & $0$ \\
Attention dropout & Uniform $[0,0.5]$ & $0.2$ \\
FFN dropout & Uniform $[0,0.5]$ & $0.1$ \\
FFN factor & Uniform $[2 / 3,8 / 3]$ & $4 / 3$\\
Learning rate & LogUniform$[1e-5,1e-3]$ & $1e-4$ \\
Weight decay & LogUniform $[1 e-6,1 e-3]$ &  $1e-5$\\
kv compression & [True, False] & True \\
kv compression sharing & [headwise, key-value] & headwise \\
Learning rate scheduler & [True, False] & False \\
Batch size & $[256, 512, 1024]$ & 512 \\
\hline
\end{tabular}
\caption{FT Transformer hyperparameters space}
\end{table}

\begin{table}[hbt!]
\rowcolors{2}{white}{gray!25}
\begin{tabular}{lll}
\hline Parameter & Distribution & Default \\
\hline Num layers & UniformInt $[1,16]$ & $8$ \\
Layer size & UniformInt $[64,1024]$ & $256$ \\
Hidden factor & Uniform $[1,4]$ & $2$ \\
Hidden dropout & $[0,0.5]$ & $0.2$ \\
Residual dropout & Uniform$[0,0.5]$ & $0.2$\\
Learning rate & LogUniform$[1e-5, 1 e-2]$ & $1e-3$ \\
Weight decay & LogUniform $[1e-8,1 e-3]$ & $ 1e-7$\\
Category embedding size & UniformInt $[64,512]$ & $128$ \\
Normalization & [batchnorm, layernorm] & batchnorm \\
Learning rate scheduler & [True, False] & True \\
Batch size & $[256, 512, 1024]$ & $512$ \\
\hline
\end{tabular}
\caption{Resnet hyperparameters space}
\end{table}

\begin{table}[hbt!]
\rowcolors{2}{white}{gray!25}
\begin{tabular}{lll}
\hline Parameter & Distribution & Default \\
\hline Num layers & UniformInt $[1,8]$ & $4$ \\
Layer size & UniformInt $[16,1024]$ & $256$ \\
Dropout & $[0,0.5]$ & $0.2$ \\
Learning rate & LogUniform$[1e-5, 1 e-2]$ & $1e-3$ \\
Category embedding size & UniformInt $[64,512]$ & $128$ \\
Learning rate scheduler & [True, False] & True \\
Batch size & $[256, 512, 1024]$ & $512$ \\
\hline
\end{tabular}
\caption{MLP hyperparameters space}
\end{table}

\begin{table}[hbt!]
\rowcolors{2}{white}{gray!25}
\begin{tabular}{lll}
\hline Parameter & Distribution & Default \\
\hline Num layers & UniformInt $[1, 2, 3, 6, 12]$ & $3$ \\
Num heads & $[2, 4, 8]$ & $4$\\
Layer size & UniformInt $[32, 64, 128]$ & $128$ \\
Dropout & $[0, 0.1, 0.2, 0.3, 0.4, 0.5, 0.6, 0.7, 0.8]$ & $0.1$ \\
Learning rate & LogUniform$[1e-5, 1e-3]$ & $3e-5$ \\
Batch size & $[128, 256]$ & $512$ \\
\hline
\end{tabular}
\caption{SAINT hyperparameters space}
\end{table}

\begin{table}[hbt!]
\rowcolors{2}{white}{gray!25}
\begin{tabular}{ll}
\hline Parameter & Distribution \\
\hline Max depth & UniformInt[1,11] \\
Num estimators & UniformInt[100, 6000, 200] \\
Min child weight & LogUniformInt[1, 1e2] \\
Subsample &  Uniform[0.5,1] \\
Learning rate & LogUniform[1e-5,0.7] \\
Col sample by level & Uniform[0.5,1] \\
Col sample by tree & Uniform[0.5, 1] \\
Gamma & LogUniform[1e-8,7] \\
Lambda & LogUniform[1,4] \\
Alpha & LogUniform[1e-8,1e2] \\
\hline
\end{tabular}
\caption{XGBoost hyperparameters space}
\end{table}

\begin{table}[hbt!]
\rowcolors{2}{white}{gray!25}
\begin{tabular}{ll}
\hline Parameter & Distribution \\
\hline Max depth & [None, 2, 3, 4] ([0.7, 0.1, 0.1, 0.1]) \\
Num estimators & LogUniformInt[9.5, 3000.5] \\
Criterion & [gini, entropy] (classif) [squared\_error, absolute\_error] (regression) \\
Max features &  [sqrt, sqrt, log2, None, 0.1, 0.2, 0.3, 0.4, 0.5, 0.6, 0.7, 0.8, 0.9] \\
Min samples split & [2, 3] ([0.95, 0.05]) \\
Min samples leaf & LogUniformInt[1.5, 50.5] \\
Boostrap & [True, False] \\
Min impurity decrease & [0.0, 0.01, 0.02, 0.05] ([0.85, 0.05, 0.05, 0.05])\\
\hline
\end{tabular}
\caption{RandomForest hyperparameters space}
\end{table}

\begin{table}[hbt!]
\rowcolors{2}{white}{gray!25}
\begin{tabular}{lp{9cm}}
\hline Parameter & Distribution \\
\hline Loss & [deviance, exponential] (classif) [squared\_error, absolute\_error, huber] (regression) \\
Learning rate & LogNormal[log(0.01), log(10)] \\
Subsample & Uniform[0.5, 1] \\
Num estimators & LogUniformInt[10.5, 1000.5] \\
Criterion & [friedman\_mse, squared\_error] \\
Max depth & [None, 2, 3, 4, 5] ([0.1, 0.1, 0.6, 0.1, 0.1]) \\
Min samples split & [2, 3] ([0.95, 0.05]) \\
Min samples leaf & LogUniformInt[1.5, 50.5] \\
Min impurity decrease & [0.0, 0.01, 0.02, 0.05] ([0.85, 0.05, 0.05, 0.05]) \\
Max leaf nodes & [None, 5, 10, 15] ([0.85, 0.05, 0.05, 0.05]) \\
\hline
\end{tabular}
\caption{GradientBoosting hyperparameters space}
\end{table}

\begin{table}[hbt!]
\rowcolors{2}{white}{gray!25}
\begin{tabular}{ll}
\hline Parameter & Distribution \\
\hline Loss & [squared\_error, absolute\_error] (regression) \\
Learning rate & LogNormal[log(0.01), log(10)] \\
Max depth & [None, 2, 3, 4] ([0.1, 0.1, 0.7, 0.1]) \\
Min samples leaf & NormalInt[20, 2] \\
Max leaf nodes & NormalInt[31, 5] \\
\hline
\end{tabular}
\caption{HistGradientBoosting hyperparameters space}
\end{table}

\clearpage

\paragraph{Other details} To run the random searches, we use the "sweep" functionality of Weight and Biases \citep{biewaldExperimentTrackingWeights2020}.

\paragraph{Dataset by dataset}

\begin{figure}[hbt!]
    \centering
    \includegraphics[width=\linewidth]{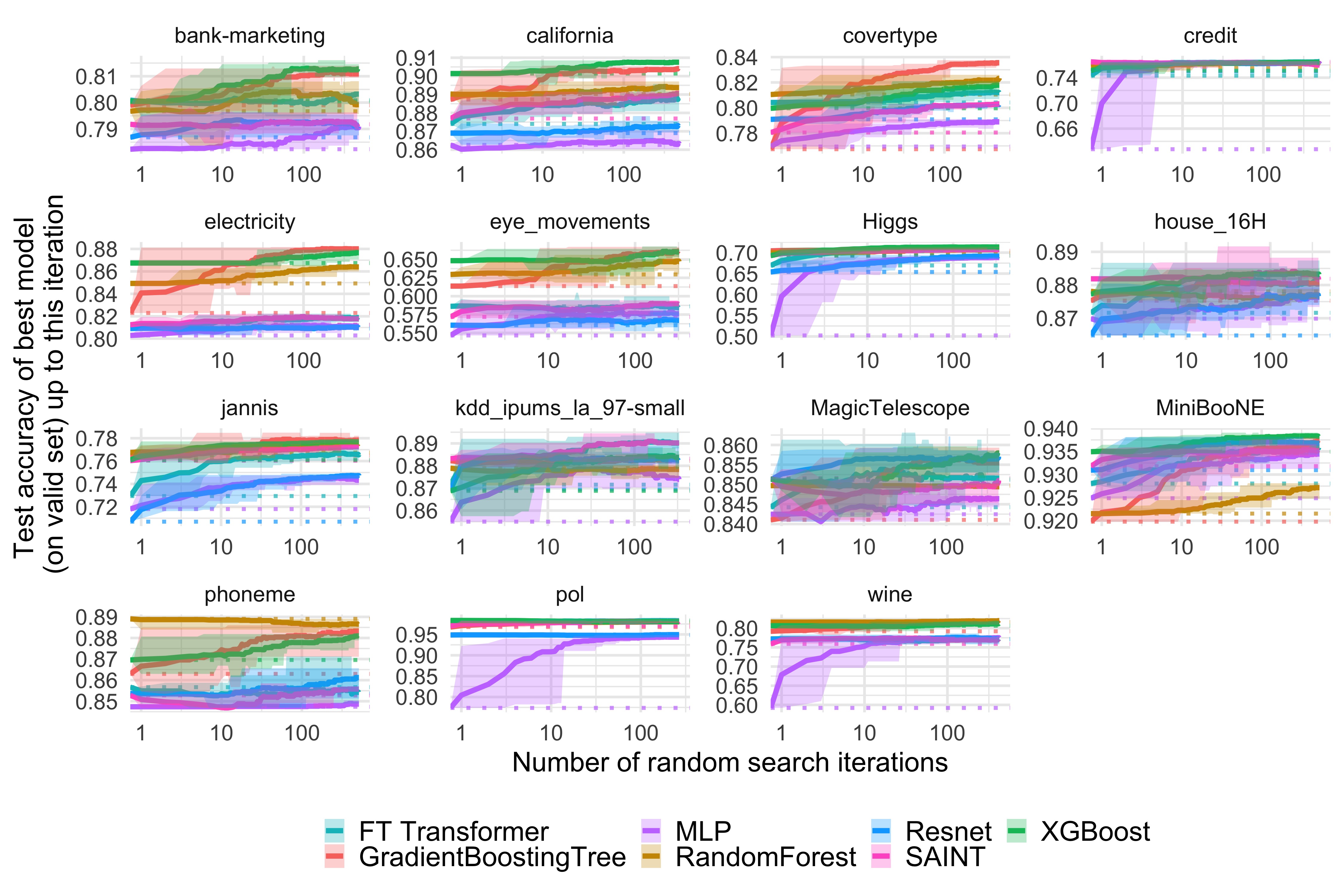}
    \caption{\textbf{Unormalized benchmark results for classification tasks on numerical features only.} Medium-sized setting.}
    \label{fig:benchmark_classif_numerical_datasets}
\end{figure}

\begin{figure}[hbt!]
    \centering
    \includegraphics[width=\linewidth]{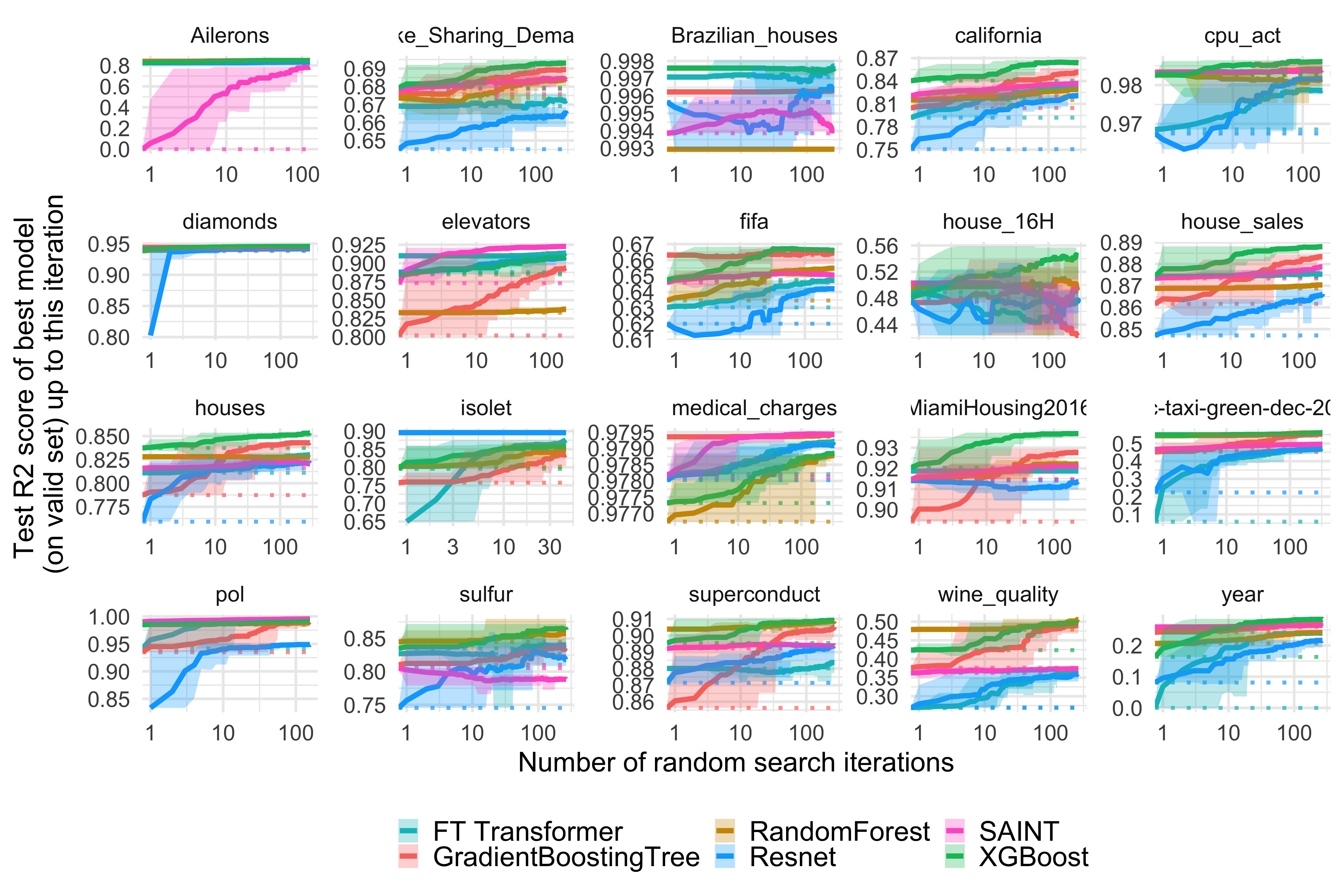}
    \caption{\textbf{Unormalized benchmark results for regression tasks on numerical features only.} Medium-sized setting. Negative values are truncated to zero to make the plots easier to read.}
    \label{fig:benchmark_regression_numerical_datasets}
\end{figure}

\begin{figure}[hbt!]
    \centering
    \includegraphics[width=\linewidth]{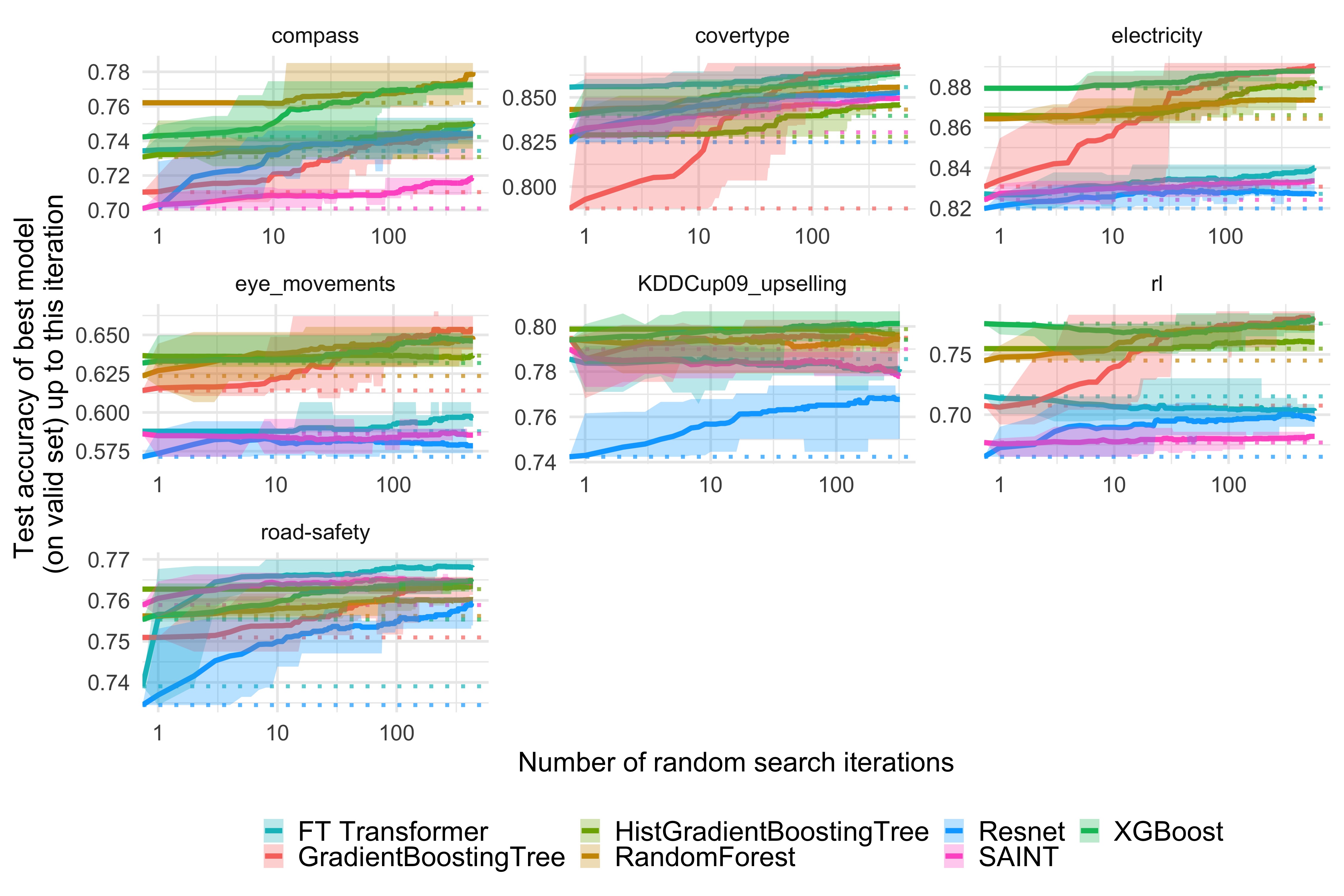}
    \caption{\textbf{Unormalized benchmark results for classification tasks on both categorical and numerical features.} Medium-sized setting.}
    \label{fig:benchmark_categorical_classif_datasets}
\end{figure}

\begin{figure}[hbt!]
    \centering
    \includegraphics[width=\linewidth]{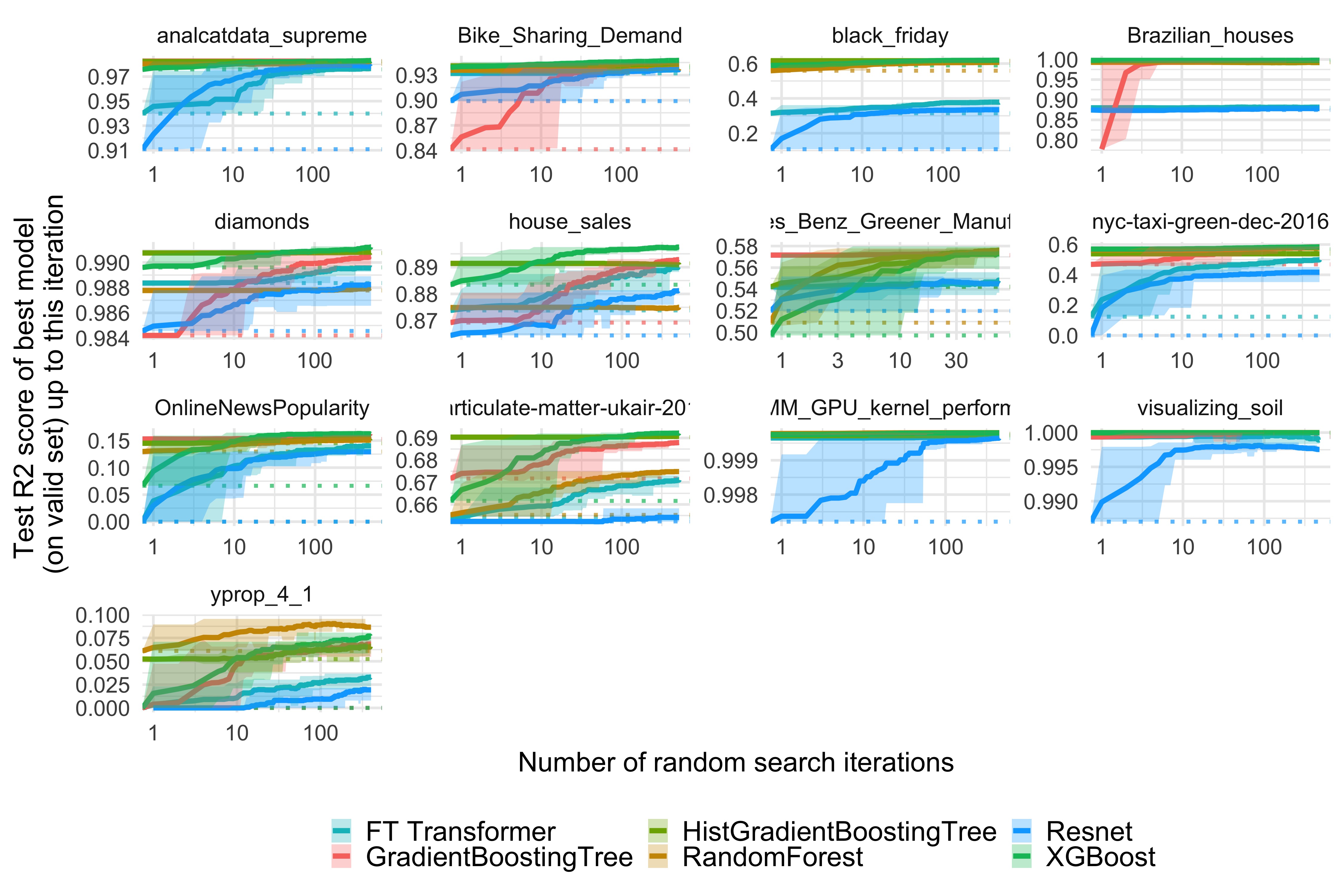}
    \caption{\textbf{Unormalized benchmark results for regression tasks on both categorical and numerical features.} Medium-sized setting. Negative values are truncated to zero to make the plots easier to read.}
    \label{fig:benchmark_categorical_regression_datasets}
\end{figure}

\begin{figure}[hbt!]
\begin{subfigure}{.5\textwidth}
  
  \includegraphics[width=\linewidth]{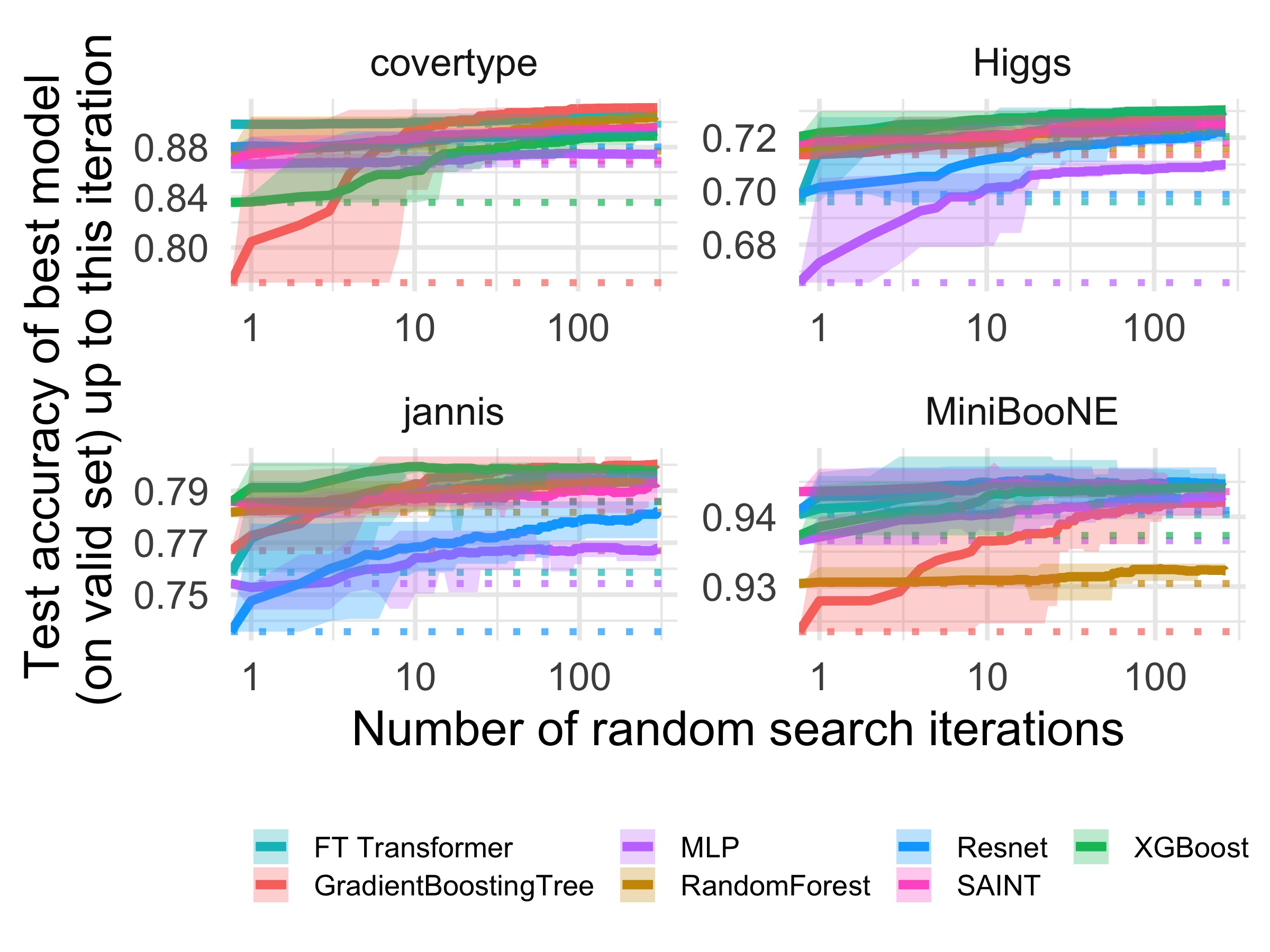}%
        \llap{\raisebox{.8\linewidth}{\parbox{.8\linewidth}{\sffamily{\bfseries Classification}}}}%
\end{subfigure}%
\begin{subfigure}{.5\textwidth}
  \includegraphics[width=\linewidth]{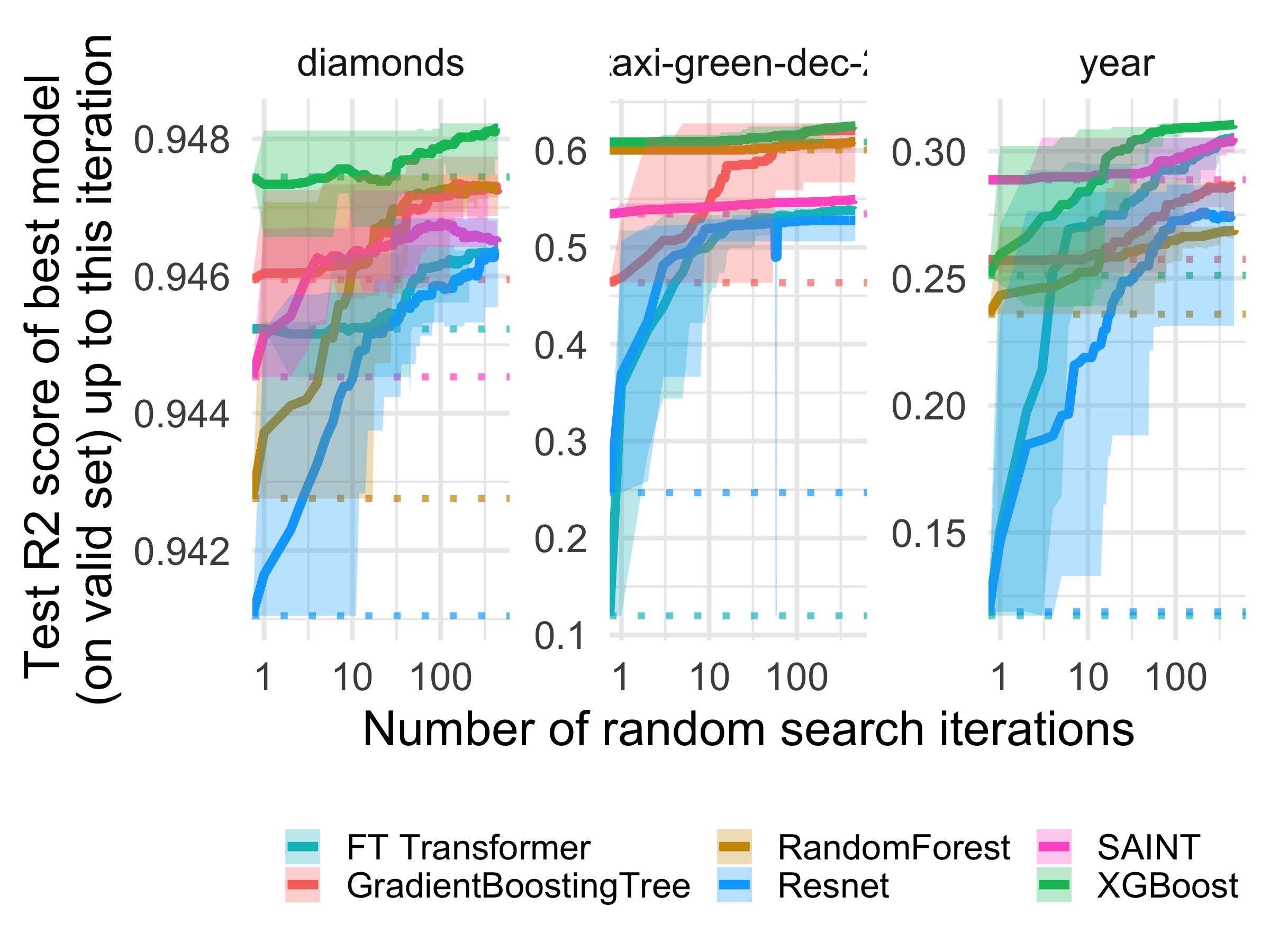}%
          \llap{\raisebox{.8\linewidth}{\parbox{.8\linewidth}{\sffamily{\bfseries Regression}}}}%
\end{subfigure}
\caption{\textbf{Unormalized benchmark results for large scale datasets, for numerical features only.}}
\label{fig:benchmark_large_datasets_numerical}

\bigskip

\begin{subfigure}{.5\textwidth}
  \includegraphics[width=\linewidth]{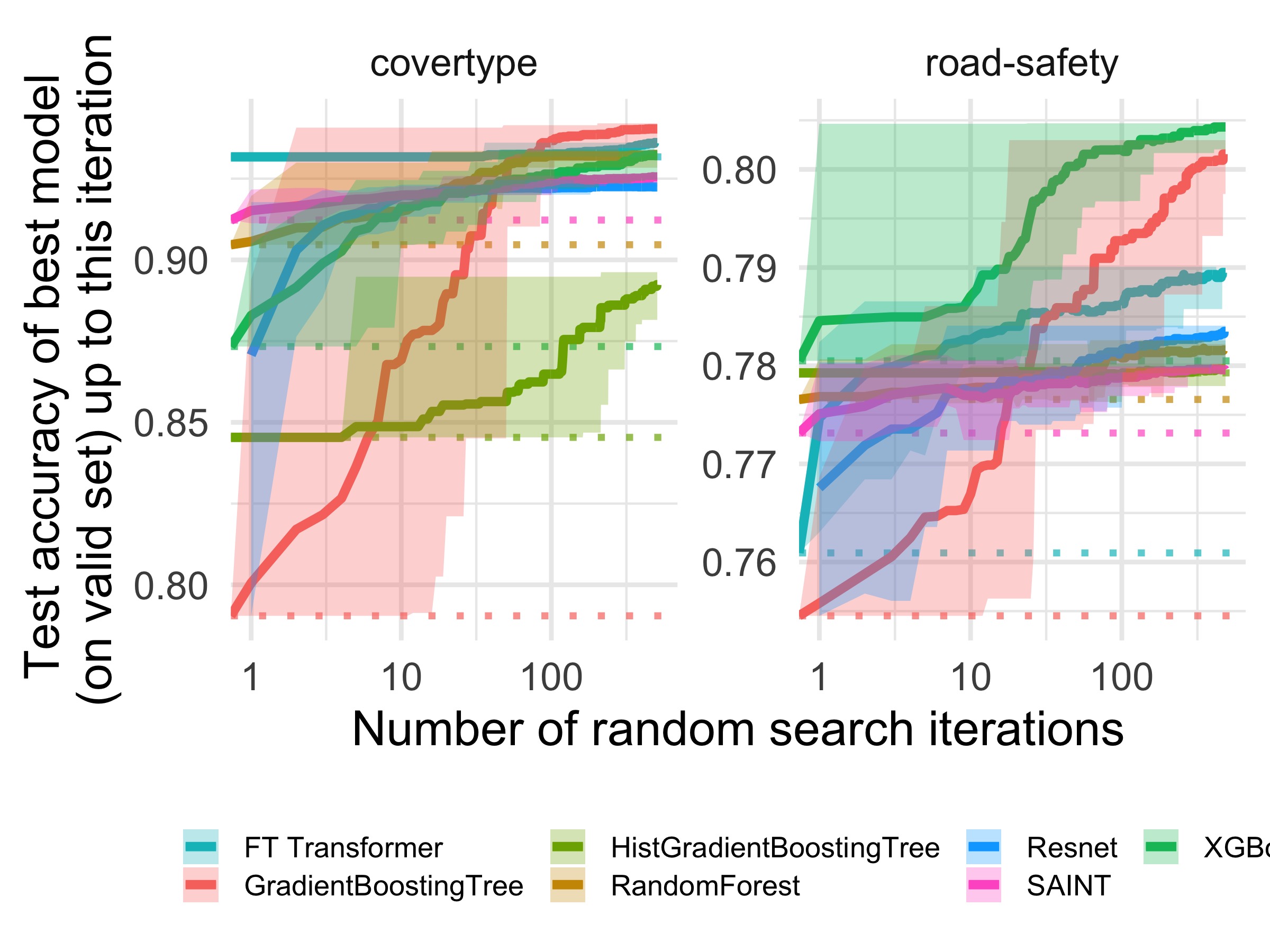}%
        \llap{\raisebox{.8\linewidth}{\parbox{.8\linewidth}{\sffamily{\bfseries Classification}}}}%
\end{subfigure}%
\begin{subfigure}{.5\textwidth}
  \includegraphics[width=\linewidth]{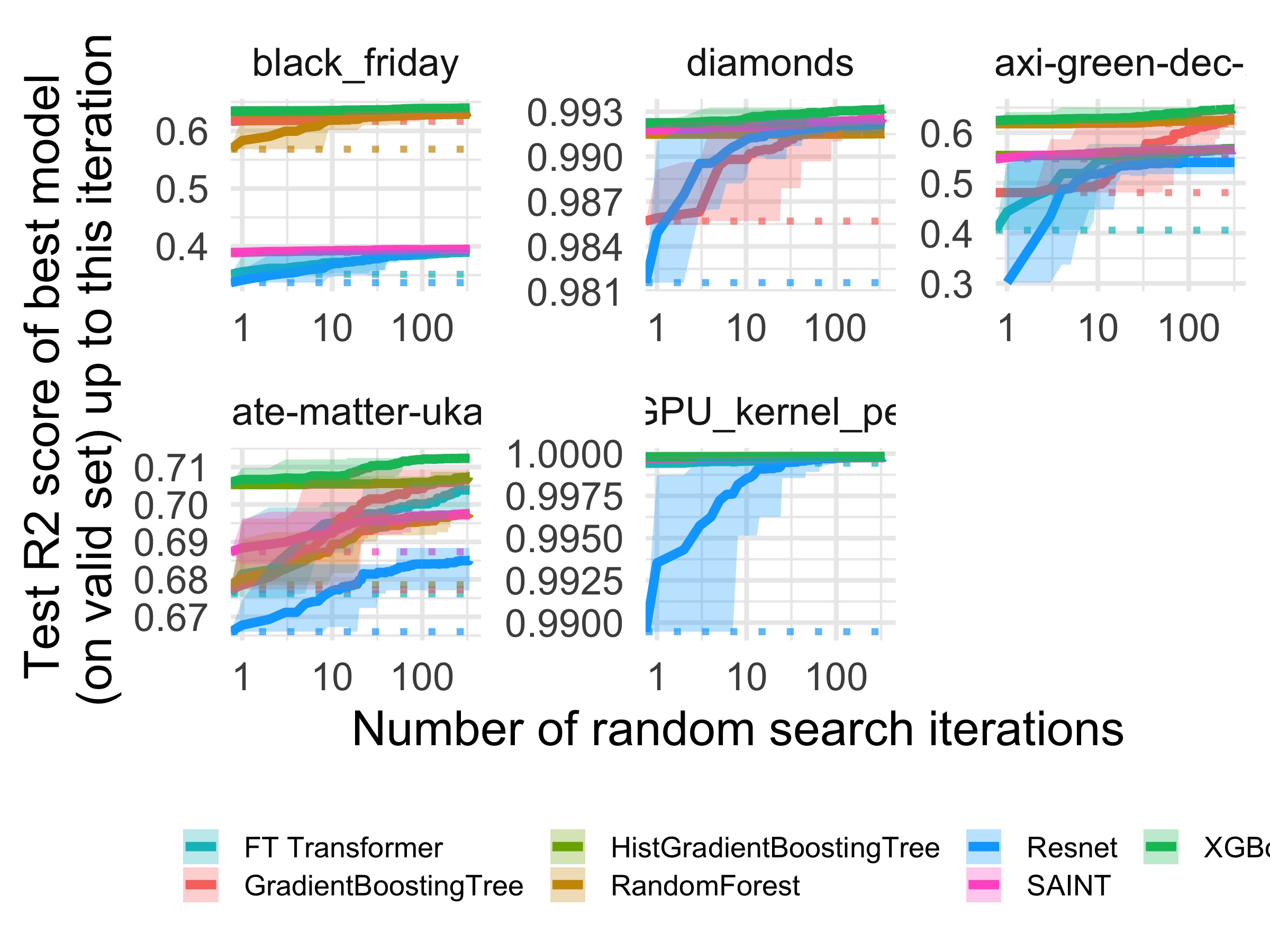}%
          \llap{\raisebox{.8\linewidth}{\parbox{.8\linewidth}{\sffamily{\bfseries Regression}}}}%
\end{subfigure}
\caption{\textbf{Unormalized benchmark results for large-scale datasets, for both numerical and categorical features.}}
\label{fig:benchmark_large_datasets_categorical}
\end{figure}

We show all unnormalized benchmark results dataset by dataset: Figure \ref{fig:benchmark_classif_numerical_datasets}, \ref{fig:benchmark_regression_numerical_datasets}, \ref{fig:benchmark_categorical_classif_datasets} and \ref{fig:benchmark_categorical_regression_datasets} for the medium-size setting, and Figure \ref{fig:benchmark_large_datasets_numerical} and \ref{fig:benchmark_large_datasets_categorical} for the large-size setting.

\cleardoublepage

\subsection{More details on experiments}\label{supp:xp_details}
In this section, we give more details on the choices we made when
creating our experiments. As results may sometimes be easier to interpret
before aggregation across datasets, we also show the results of each experiment on each dataset.

\subsubsection{Finding 1: NNs are biased to overly smooth solutions}

\paragraph{Smoothing} We use a Gaussian smoothing Kernel
$$
K(\mathbf{x^*}, \mathbf{x}) = \exp \left(-\frac{1}{2}(\mathbf{x^*}-\mathbf{x})^{\mathrm{T}} \mathbf{\Sigma}^{-1}(\mathbf{x^*}-\mathbf{x})\right)
$$
with $\mathbf{\Sigma}$ the estimated empirical covariance multiplied by the "squared lengthscale". The transformed target on the train set becomes:
$$
\tilde{Y}\left(X_{i}\right)=\frac{\sum_{j=1}^{N} K\left(X_{i}, X_{j}\right) Y\left(X_{j}\right)}{\sum_{j=1}^{N} K\left(X_{i}, X_{j}\right)}
$$
with $(X_1, ..X_N)$ the training set covariates and $(Y_1, ..., Y_N)$ the training set original targets.

\paragraph{More details}

\begin{itemize}
    \item We restrict all datasets to their 5 most important features (according to a RandomForest feature importance ranking). This makes the smoothing easier, as kernel smoothing can be hard in high-dimension, while keeping enough features to produce interesting results.
    \item We estimate the covariance matrix of these features through ScikitLearn's {\tt MinCovDet}, which is more robust to outliers than the empirical covariance.
\end{itemize}

\begin{figure}[!htb]
    \centering
    \includegraphics[width=\linewidth]{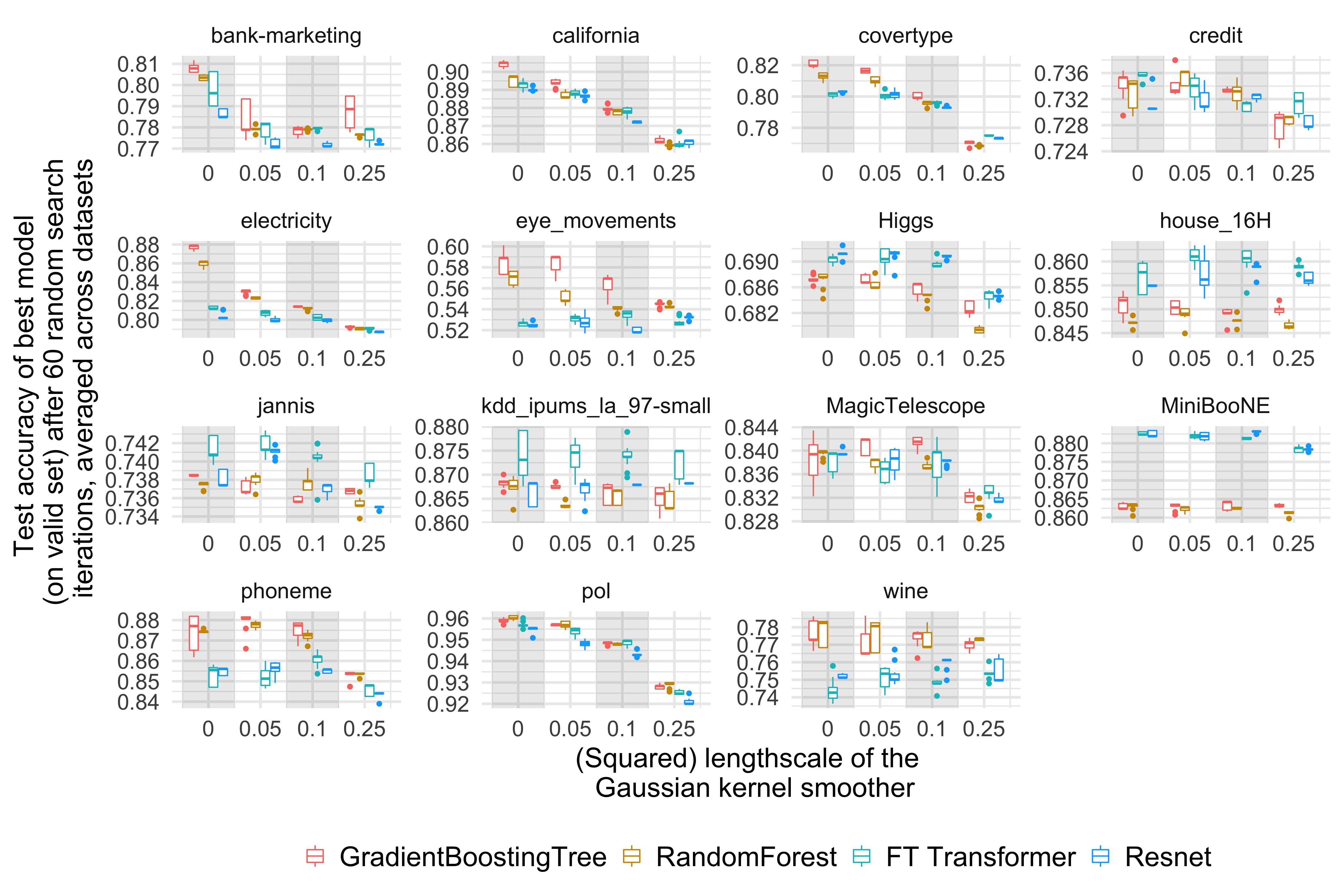}
    \caption{\textbf{Test accuracy of different models for varying smoothing of the target function on the train set}. We smooth the target function through a Gaussian Kernel smoother, whose covariance matrix is the data covariance, multiplied by the (squared) lengthscale of the Gaussian kernel smoother. A lengthscale of 0 corresponds to no smoothing (the original data). All features have been Gaussienized before the smoothing through ScikitLearn's QuantileTransformer. The boxplots represent the distribution of accuracies across 15 re-orderings of the random search. Same experiment than Fig. \ref{fig:high_frequencies}, shown for each dataset without score normalization}
    \label{fig:high_frequency_dataset}
\end{figure}

Raw results are shown in Figure \ref{fig:high_frequency_dataset} dataset by dataset.

\paragraph{Examples of irregular patterns}

\begin{figure}[hbt!]
\begin{subfigure}{.5\textwidth}
  
  \includegraphics[width=\linewidth]{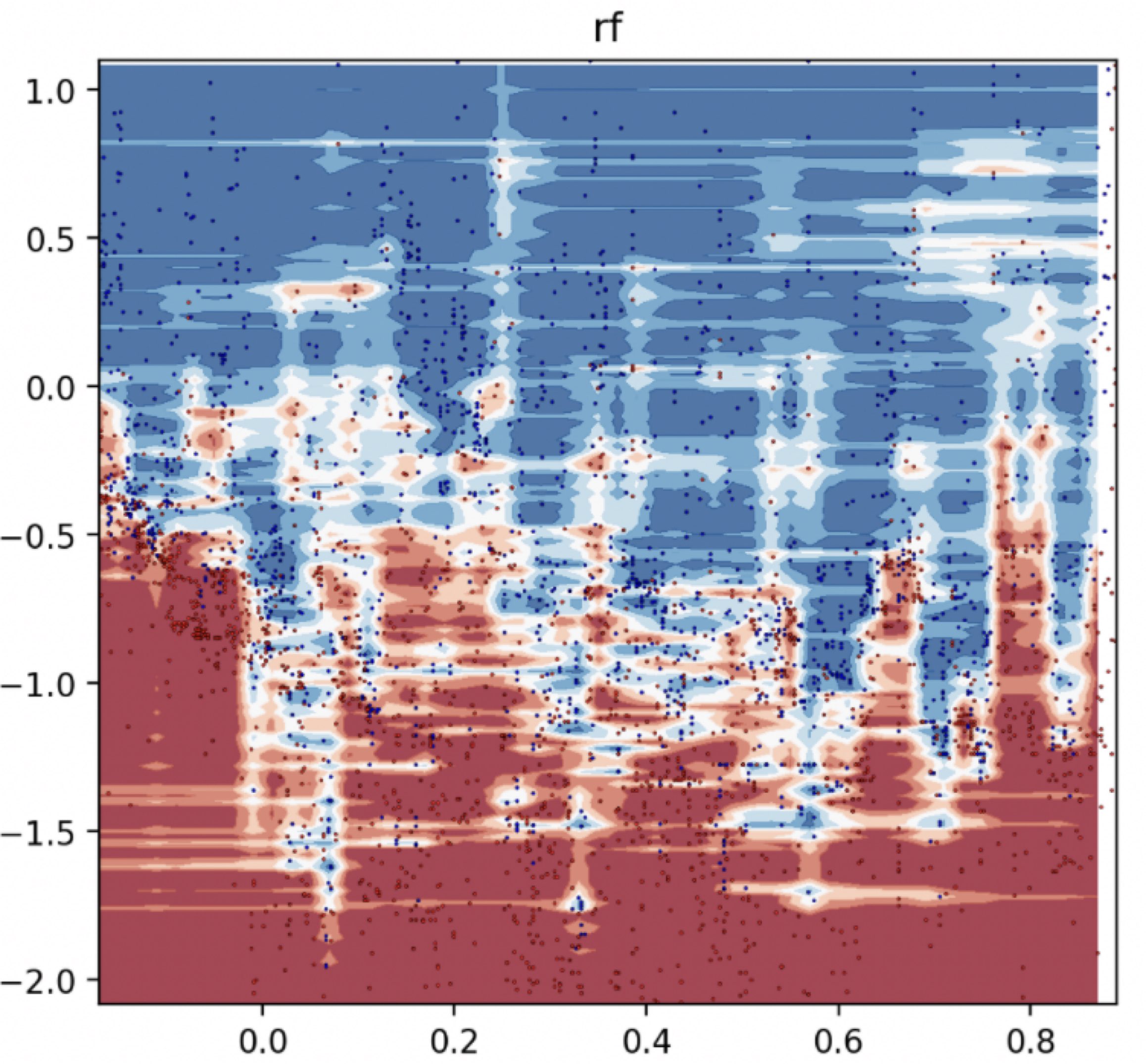}%
        \llap{\raisebox{.8\linewidth}{\parbox{.8\linewidth}{\sffamily{\bfseries RandomForest (85\%)}}}}%
\end{subfigure}%
\begin{subfigure}{.5\textwidth}
  \includegraphics[width=\linewidth]{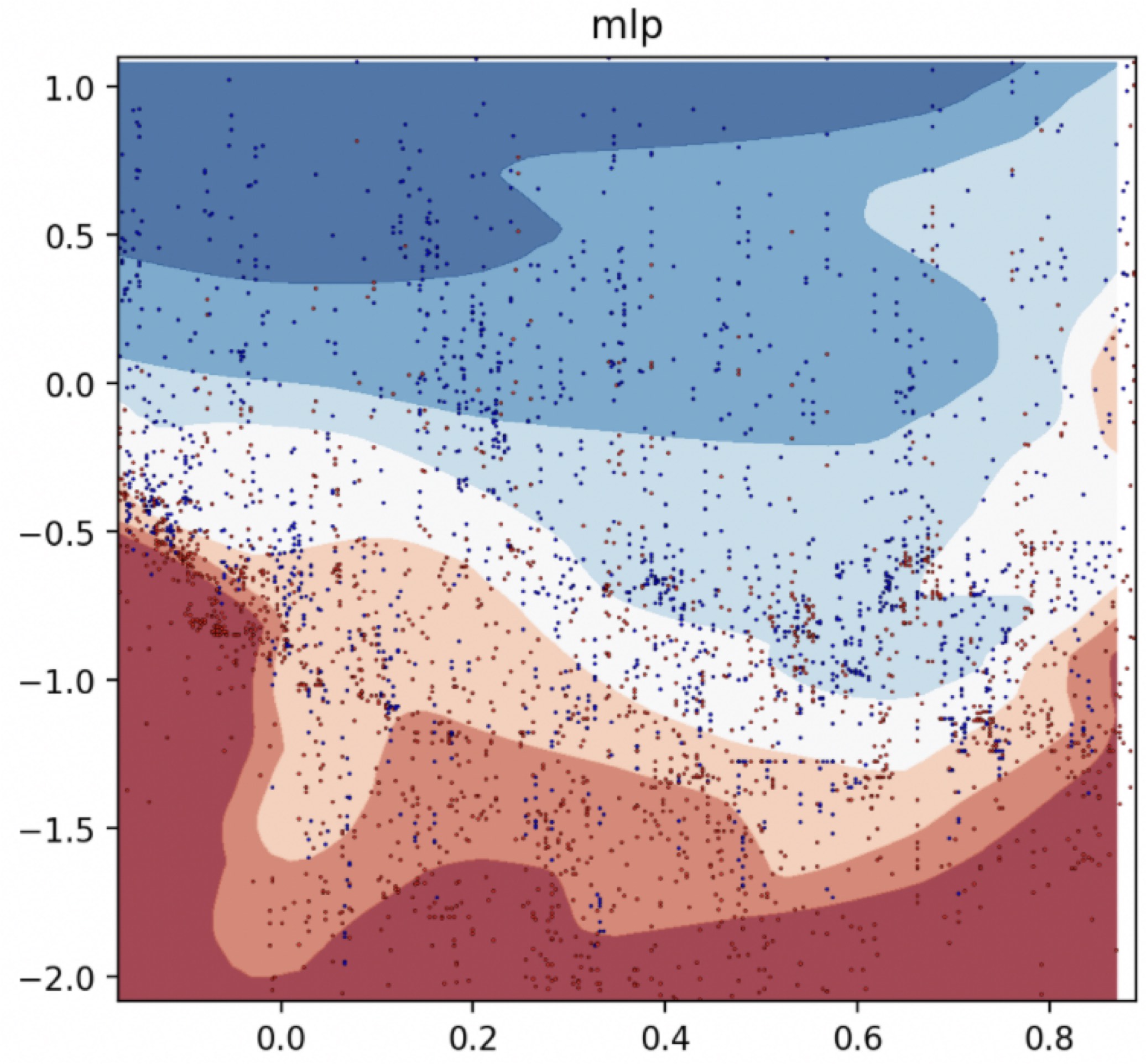}%
          \llap{\raisebox{.8\linewidth}{\parbox{.8\linewidth}{\sffamily{\bfseries MLP (80\%)}}}}%
\end{subfigure}
\caption{Decision boundaries of a default MLP and RandomForest for the 2 most important features of the \textit{electricity} dataset}
\label{fig:irregular_pattern}
\end{figure}

Figure \ref{fig:irregular_pattern} shows the decision boundaries of a default MLP and a default RandomForest on the 2 most important features of the \textit{electricity} dataset. The RandomForest achieve a perfect training accuracy and a test accuracy (85\%) higher than the MLP (80\%). The features are Gaussienied and we show a zoomed-in part of the feature space. In this part, we can see that the RandomForest is able to learn irregular patterns on the x axis (which corresponds to the \textit{date} feature) that the MLP does not learn. We show this difference for default hyperparameters but it seems to us that this is a typical behavior of neural networks, and it is actually hard, albeit not impossible, to find hyperparameters to successfully learn these patterns. 

\subsubsection{Finding 2: Uninformative features affect more MLP-like NNs}

\paragraph{Tabular datasets contain a lot of uninformative features}

\begin{figure}[!htb]
    \centering
    \includegraphics[width=\linewidth]{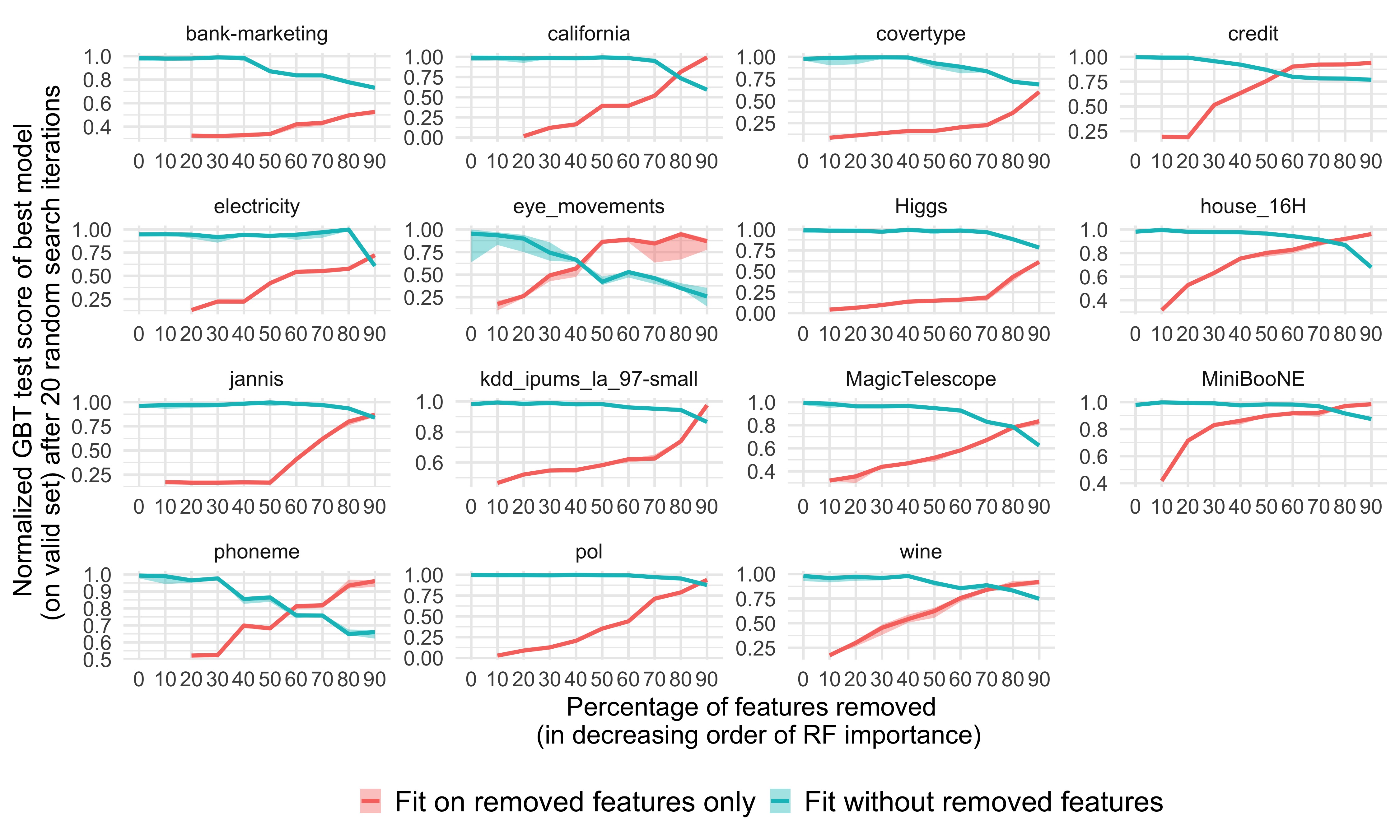}
    \caption{\textbf{Test accuracy of a GBT for varying proportions of removed features}. Features are removed in increasing order of feature importance (computed with a Random Forest), and the two lines correspond to the accuracy using the (most important) kept features (blue) or the (least important) removed features (red). Scores are normalized between 0 (random chance) and 1 (best score among all hyperparameters). These scores are averaged across 30 random search orders, and the ribbons correspond to the minimum and maximum values among these 30 orders. Same experiment than Fig. \ref{fig:removed_features}, shown for each dataset.  Note that axes do not always start at zero.}
    \label{fig:remove_features_datasets}
\end{figure}

Raw results are shown in Figure \ref{fig:remove_features_datasets} dataset by dataset.

\paragraph{Uninformative features affect more MLP-like NNs}

\begin{figure}[!htb]
    \centering
    \includegraphics[width=\linewidth]{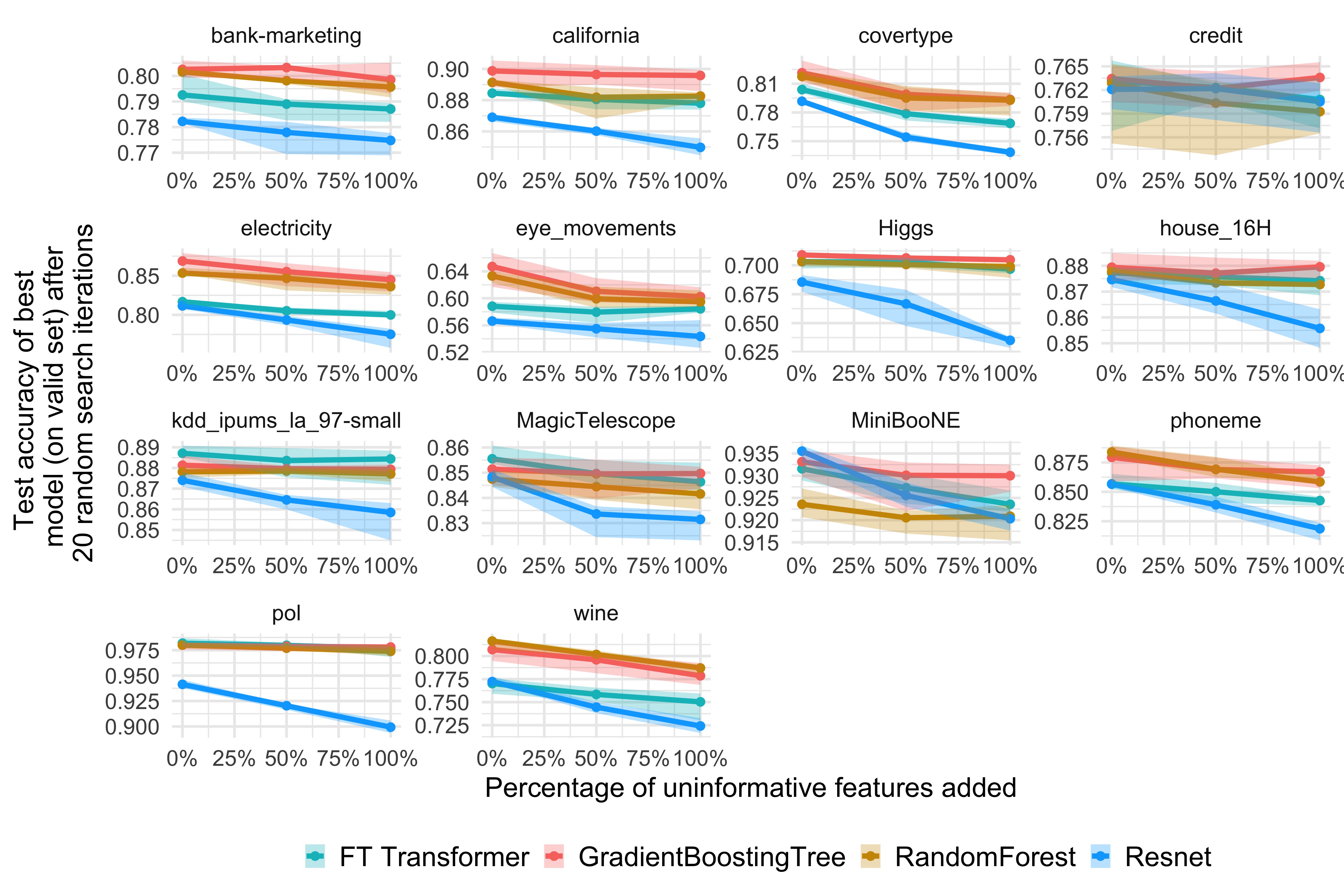}
    \caption{\textbf{Test accuracy changes when adding uninformative features}. Added features are sampled from standard Gaussians uncorrelated with the target and with other features. Ribbons correspond to the minimum and maximum score among the 30 different random search reorders (starting with the default models). Same experiment that in Figure \ref{fig:features} (b), shown for each dataset without score normalization.}
    \label{fig:add_features_datasets}
\end{figure}

\begin{figure}[!htb]
    \centering
    \includegraphics[width=\linewidth]{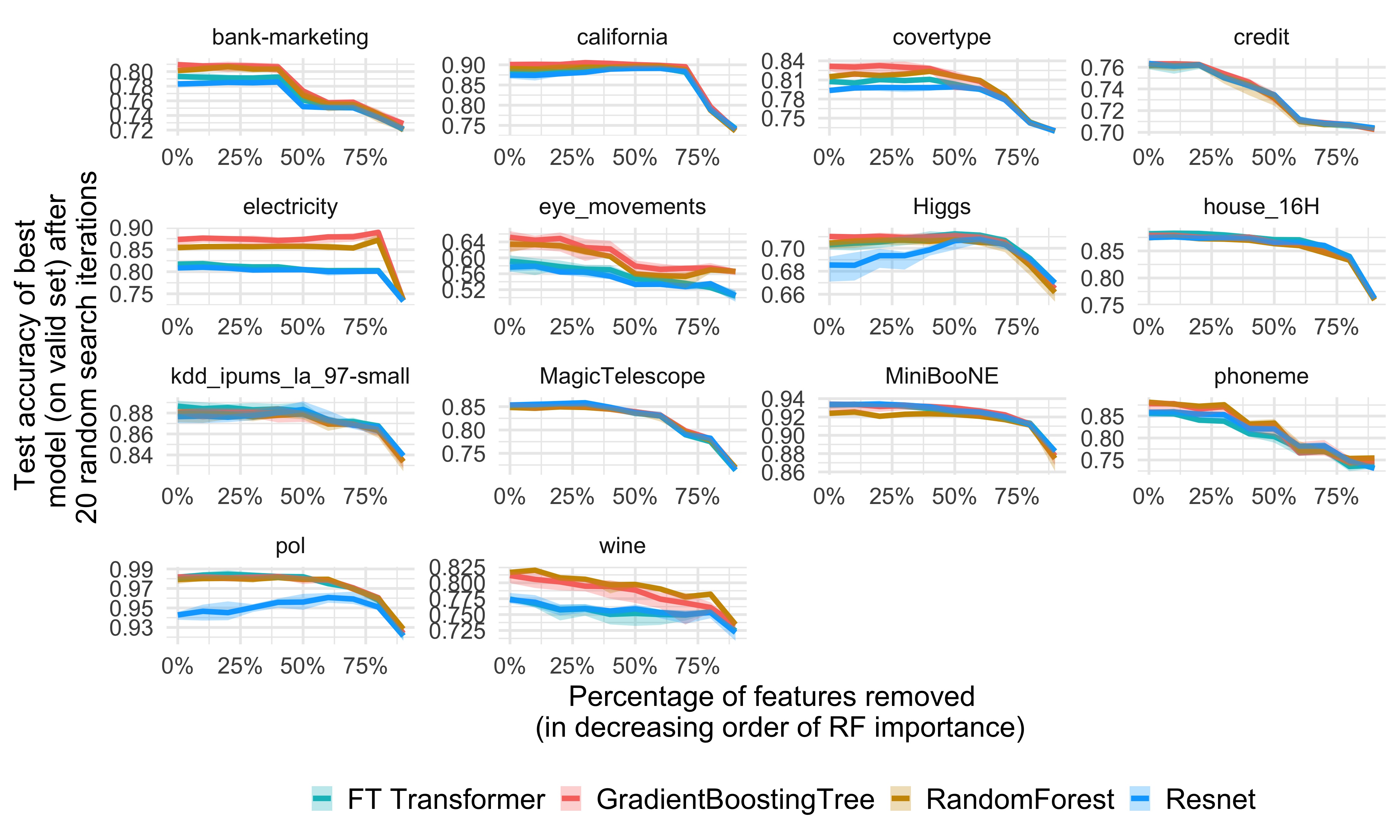}
    \caption{\textbf{Test accuracy changes when removing uninformative features}. Features are removed in increasing order of feature
importance (computed with a Random Forest). Ribbons correspond to the minimum and maximum score among the 30 different random search reorders (starting with the default models).  Same experiment that in Figure \ref{fig:features} (a), shown for each dataset without score normalization.}
    \label{fig:useless_features_datasets}
\end{figure}

Raw results are shown in Figure \ref{fig:add_features_datasets} (uninformative features added) and Figure \ref{fig:useless_features_datasets} (uninformative features removed).

\subsubsection{Finding 3: Data are non invariant by rotation, so should be learning procedures}

\paragraph{Details} Random rotation were computed using Scipy's \citep{virtanenSciPyFundamentalAlgorithms2020} \texttt{stats.special\_ortho\_group.rvs}.

\begin{figure}[!htb]
    \centering
    \includegraphics[width=\linewidth]{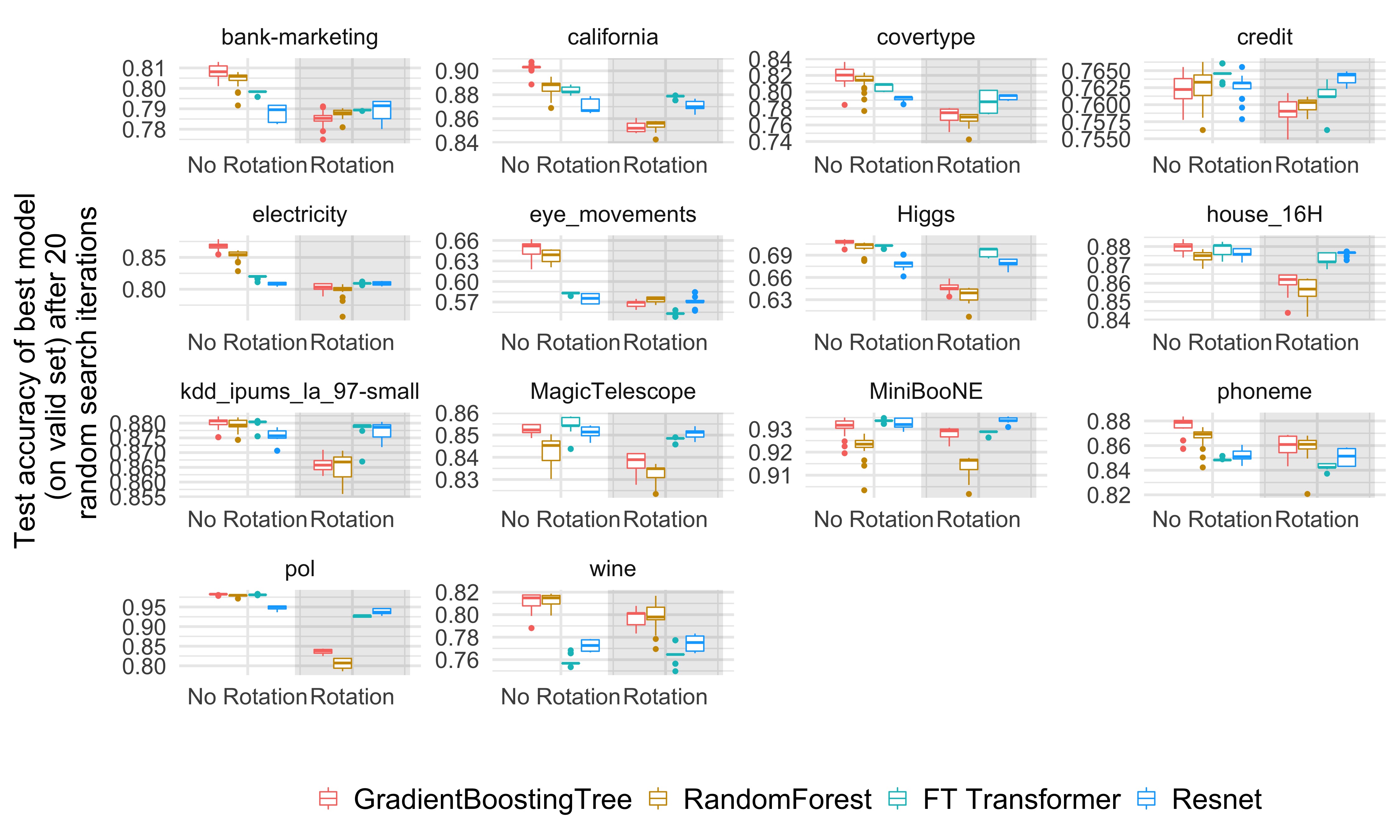}
    \caption{\textbf{Test accuracy of different models when randomly rotating our datasets}. All features are Gaussianized before the random rotations. The scores are averaged across datasets, and the boxes depict the distribution across random search shuffles. Same experiment that in \ref{fig:rotation} (Left), shown for each dataset without score normalization.}
    \label{fig:random_rotation_datasets}
\end{figure}

\begin{figure}[!htb]
    \centering
    \includegraphics[width=\linewidth]{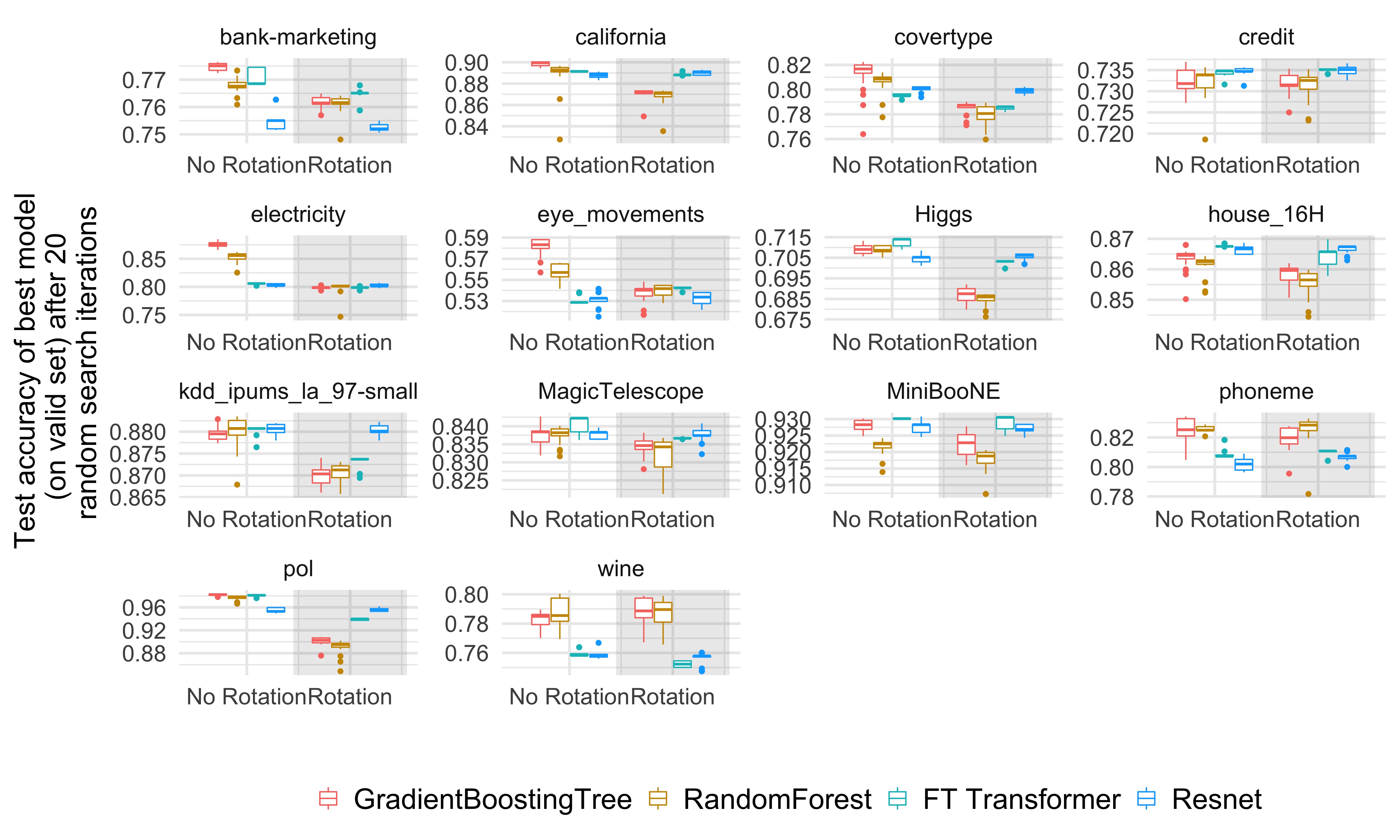}
    \caption{\textbf{Test accuracy of different models when randomly rotating our datasets, with 50\% features removed}. All features are Gaussianized before the random rotations. The removed features are the least important half (according to a RandomForest), and are removed before the rotation. The scores are averaged across datasets, and the boxes depict the distribution across random search shuffles. Same experiment that in \ref{fig:rotation} (Right), shown for each dataset without score normalization.}
    \label{fig:random_rotation_datasets_features_removed}
\end{figure}

Raw results are shown in Figure \ref{fig:random_rotation_datasets} (with all features) and Figure \ref{fig:random_rotation_datasets_features_removed} (with 50\% features) dataset by dataset.

\clearpage

\subsection{Discussion on \cite{kadraRegularizationAllYou2021}}\label{supp:search_space_reg}

We observed that tree-based models are superior for every random search
budget, and the performance gap stays wide even after a large number of
random search iteration. However, this might no longer be true when
adding additional regularization techniques to our random search, such as
data augmentation. Indeed, \cite{kadraWelltunedSimpleNets2021} find that
searching through a "cocktail" of regularization on a
Multi-Layer-Perceptron is competitive with XGBoost after half an hour of
tuning (for both models), though the datasets considered in their paper
are quite different, in particular with the presence of "deterministic"
game-inspired datasets in \cite{kadraRegularizationAllYou2021}, on which
their method performs very well and contributes markedly to the overall
benchmark results.

\subsection{Which hyperparameters perform well on tabular data?}\label{supp:search_space}

Our random search results provide insights into which hyperparameters are important for learning on tabular data. Below we present a measure of hyperparameters importance in our classification benchmark on numerical features.

\paragraph{Methodology} Accuracy is normalized (see \ref{bench}), and negative scores are truncated to zero. For each model, we fit a RandomForest classifier with default hyperparameters to predict these scores from the model's hyperparameters and the dataset, and we compute the feature importance for each of the hyperparameter ("rf\_importance"). This gives us a score which represent how much an hyperparameter should be tuned. To measure if an hyperparameter has a positive or negative impact on the performance, we also compute the coefficient of a LinearRegression trained on the same task that the RandomForest ("lin\_coef")

\paragraph{Results} The learning rate is by far the most important parameter for neural networks and gradient-boosted trees. Note that the linear coefficient is not always very high, which suggests that the learning rate should be tuned for each dataset. For tree-based models, the depth of the trees is another very important parameter, and it seems that deeper trees help, even for gradient-boosted trees. This observation is related to our "Finding 1" \ref{finding_1}, as deeper trees enable very irregular patterns to be learned.

Below we give the results for each architecture.

\paragraph{MLP} ~

{\small%
\rowcolors{2}{white}{gray!25}
\begin{filecontents*}{hp_csv/MLP_classifiation_medium_numerical_hp.csv}
names,rf_importance,lin_coef
lr,0.22,-0.04
d_layers,0.18,-0.01
d_embedding,0.15,0.0
n_layers,0.14,-0.1
batch_size,0.03,-0.01
lr_scheduler,0.01,0.0
\end{filecontents*}
\csvautobooktabular[respect underscore=true]{hp_csv/MLP_classifiation_medium_numerical_hp.csv}%
}

\paragraph{Resnet} ~

{\small%
\rowcolors{2}{white}{gray!25}
\begin{filecontents*}{hp_csv/Resnet_classifiation_medium_numerical_hp.csv}
names,rf_importance,lin_coef
lr,0.08,0.01
normalization_layernorm,0.05,0.0
n_layers,0.04,-0.03
d,0.02,0.0
hidden_dropout,0.02,0.01
residual_dropout,0.01,0.0
batch_size,0.01,0.0
optimizer__weight_decay,0.0,0.0
d_hidden_factor,0.0,0.0
d_embedding,0.0,0.0
lr_scheduler,0.0,0.0
\end{filecontents*}
\csvautobooktabular[respect underscore=true]{hp_csv/Resnet_classifiation_medium_numerical_hp.csv}%
}

\paragraph{FT Transformer} ~

{\small%
\rowcolors{2}{white}{gray!25}
\begin{filecontents*}{hp_csv/FTTransformer_classifiation_medium_numerical_hp.csv}
names,rf_importance,lin_coef
lr,0.06,0.0
residual_dropout,0.04,-0.02
d_token,0.03,0.0
n_layers,0.02,0.0
ffn_dropout,0.02,0.0
d_ffn_factor,0.01,0.0
kv_compression,0.01,-0.01
optimizer__weight_decay,0.01,0.0
attention_dropout,0.01,0.0
lr_scheduler,0.0,0.0
kv_compression_sharing_key-value,0.0,0.0
batch_size,0.0,0.0
\end{filecontents*}
\csvautobooktabular[respect underscore=true]{hp_csv/FTTransformer_classifiation_medium_numerical_hp.csv}%
}

\paragraph{SAINT} ~

{\small%
\rowcolors{2}{white}{gray!25}
\begin{filecontents*}{hp_csv/SAINT_classifiation_medium_numerical_hp.csv}
names,rf_importance,lin_coef
lr,0.28,-0.11
depth,0.15,-0.07
dim,0.05,-0.03
dropout,0.03,0.0
heads,0.01,0.0
batch_size,0.01,0.0
val_batch_size,0.0,0.0
\end{filecontents*}
\csvautobooktabular[respect underscore=true]{hp_csv/SAINT_classifiation_medium_numerical_hp.csv}%
}

\paragraph{XGBoost} ~

{\small%
\rowcolors{2}{white}{gray!25}
\begin{filecontents*}{hp_csv/XGBoost_classifiation_medium_numerical_hp.csv}
names,rf_importance,lin_coef
learning_rate,0.26,0.01
min_child_weight,0.07,-0.04
max_depth_2.0,0.05,0.13
reg_alpha,0.05,-0.03
n_estimators,0.03,0.01
max_depth_4.0,0.02,0.27
subsample,0.02,0.01
colsample_bytree,0.02,0.0
max_depth_3.0,0.02,0.21
reg_lambda,0.01,0.0
gamma,0.01,0.0
max_depth_10.0,0.01,0.35
max_depth_11.0,0.01,0.35
max_depth_7.0,0.01,0.33
max_depth_5.0,0.01,0.3
max_depth_6.0,0.01,0.32
max_depth_8.0,0.01,0.34
max_depth_9.0,0.01,0.34
colsample_bylevel,0.01,0.0
\end{filecontents*}
\csvautobooktabular[respect underscore=true]{hp_csv/XGBoost_classifiation_medium_numerical_hp.csv}%
}

\paragraph{GradientBoostingClassifier} ~

{\small%
\rowcolors{2}{white}{gray!25}
\begin{filecontents*}{hp_csv/GradientBoostingTree_classifiation_medium_numerical_hp.csv}
names,rf_importance,lin_coef
learning_rate,0.4,-0.02
n_estimators,0.15,0.09
max_depth_None,0.14,0.36
max_depth_4.0,0.04,0.19
max_depth_3.0,0.02,0.11
min_samples_leaf,0.01,0.0
subsample,0.01,0.0
loss_exponential,0.0,0.0
\end{filecontents*}
\csvautobooktabular[respect underscore=true]{hp_csv/GradientBoostingTree_classifiation_medium_numerical_hp.csv}%
}

\paragraph{RandomForest} ~

{\small%
\rowcolors{2}{white}{gray!25}
\begin{filecontents*}{hp_csv/RandomForest_classifiation_medium_numerical_hp.csv}
names,rf_importance,lin_coef
max_depth_None,0.43,0.54
max_depth_4.0,0.07,0.24
max_depth_3.0,0.03,0.13
min_samples_leaf,0.01,-0.01
n_estimators,0.01,0.0
max_features_None,0.01,-0.06
bootstrap,0.01,0.01
max_features_log2,0.0,0.04
criterion_gini,0.0,0.0
max_features_0.2,0.0,0.04
max_features_sqrt,0.0,0.04
max_features_0.4,0.0,0.05
max_features_0.5,0.0,0.04
max_features_0.6,0.0,0.04
max_features_0.7,0.0,0.03
max_features_0.8,0.0,0.01
max_features_0.9,0.0,-0.01
max_features_0.3,0.0,0.05
\end{filecontents*}
\csvautobooktabular[respect underscore=true]{hp_csv/RandomForest_classifiation_medium_numerical_hp.csv}%
}

\subsection{How to use our benchmark?}\label{supp:howto}

All instructions to use our benchmark and reproduce our results are available at our repository: \url{https://github.com/LeoGrin/tabular-benchmark}. To ease the use of our benchmark and the reproducibility of our results we provide:
\begin{itemize}
    \item The selected and transformed datasets as an OpenML suite. All links to the transformed and original datasets are also in \ref{supp:datasets}.
    \item A CSV file containing the results of all our random searches. It can be used to cheaply benchmark a new method.
    \item The code used to produce our benchmark and our experiments.
\end{itemize}

\end{document}